%% file: main.tex
\documentclass[runningheads]{llncs}

\usepackage[final]{eccv}

\usepackage{eccvabbrv}

\usepackage{graphicx}
\usepackage{booktabs}

\usepackage{algorithm}     
\usepackage{algpseudocode}  
\usepackage{amsmath}   
\usepackage{booktabs}
\usepackage{graphicx}
\usepackage{subcaption}  
\usepackage{caption}
\usepackage{float}       

\usepackage{array}
\usepackage{microtype}
\usepackage[table]{xcolor}
\usepackage{enumitem}

\newcommand{\methodname}{\texttt{POET}}

\usepackage[accsupp]{axessibility}  

\usepackage{hyperref}

\usepackage{orcidlink}
\usepackage{footnote}

\begin{document}

\title{\texttt{POET}: Preference Optimization for Enhanced Text-to-Image Generation} 





\makeatletter
\renewcommand*{\@fnsymbol}[1]{\ensuremath{\ifcase#1\or \star\or \dagger\or \ddagger\or \mathsection\or \mathparagraph\or \|\or **\or \dagger\dagger\or \ddagger\ddagger \else\@ctrerr\fi}}
\makeatother

\author{Ruibo Chen$^{1,2,}\thanks{Equal Contribution. Work done while Ruibo Chen was interning at TikTok.}$, ~Jiacheng Pan$^{1,\star}$,~Heng Huang$^{2,}$\thanks{Corresponding Authors.},~Zhenheng Yang$^{1,\dagger}$}
\institute{
$^1$TikTok, $^2$University of Maryland, College Park\\
\texttt{\{ruibo.chen,panjiacheng.666,yangzhenheng\}@bytedance.com,}\\ \texttt{heng@umd.edu}\\
}

\maketitle
\input{sec/0_abstract}    
\input{sec/1_intro}

\input{sec/2_related_work}
\input{sec/3_method_new}

\input{sec/4_exp}

\input{sec/5_ablation}
\input{sec/6_analyses}
\input{sec/7_conclusion}

\section*{Acknowledgements}
RC and HH were partially supported by NSF IIS 2347592, 2348169, DBI 2405416, CCF 2348306, CNS 2347617, RISE 2536663.

%
%
\bibliographystyle{splncs04}
\bibliography{main}

\input{sec/X_suppl}

\end{document}

%% file: sec/0_abstract.tex
\begin{abstract}
Recent advances in text-to-image (T2I) generation have achieved impressive results, yet existing models often struggle with simple or underspecified user prompts due to a distributional gap with their descriptive training captions. This frequently leads to suboptimal image-text alignment, aesthetics, and overall visual quality. To bridge this gap, we propose \methodname\ (\textbf{P}reference \textbf{O}ptimization for \textbf{E}nhanced \textbf{T}ext-to-Image generation), an automated prompt rewriting framework that leverages large language models (LLMs) to refine user inputs before feeding them into frozen T2I backbones. \methodname\ introduces a carefully designed composite reward system and an iterative Direct Preference Optimization (DPO) training pipeline, enabling the rewriter to learn model-preferred prompt structures directly from multimodal feedback without requiring costly high-quality supervised fine-tuning (SFT) data. Extensive evaluations across diverse T2I models and benchmarks show that our prompt rewriter consistently improves image-text alignment, visual quality, and aesthetics, outperforming strong baselines. Furthermore, we demonstrate strong transferability by showing that a rewriter trained on one T2I backbone generalizes effectively to others without needing to be retrained. These findings highlight that \methodname\ is an effective, robust, and practical model-agnostic strategy for improving T2I systems.

\keywords{Text-to-Image Generation \and Prompt Rewriting \and Reinforcement Learning}
\end{abstract}

%% file: sec/1_intro.tex
\section{Introduction}
\label{sec:intro}

\begin{figure}
    \centering
    \includegraphics[width=0.99\linewidth]{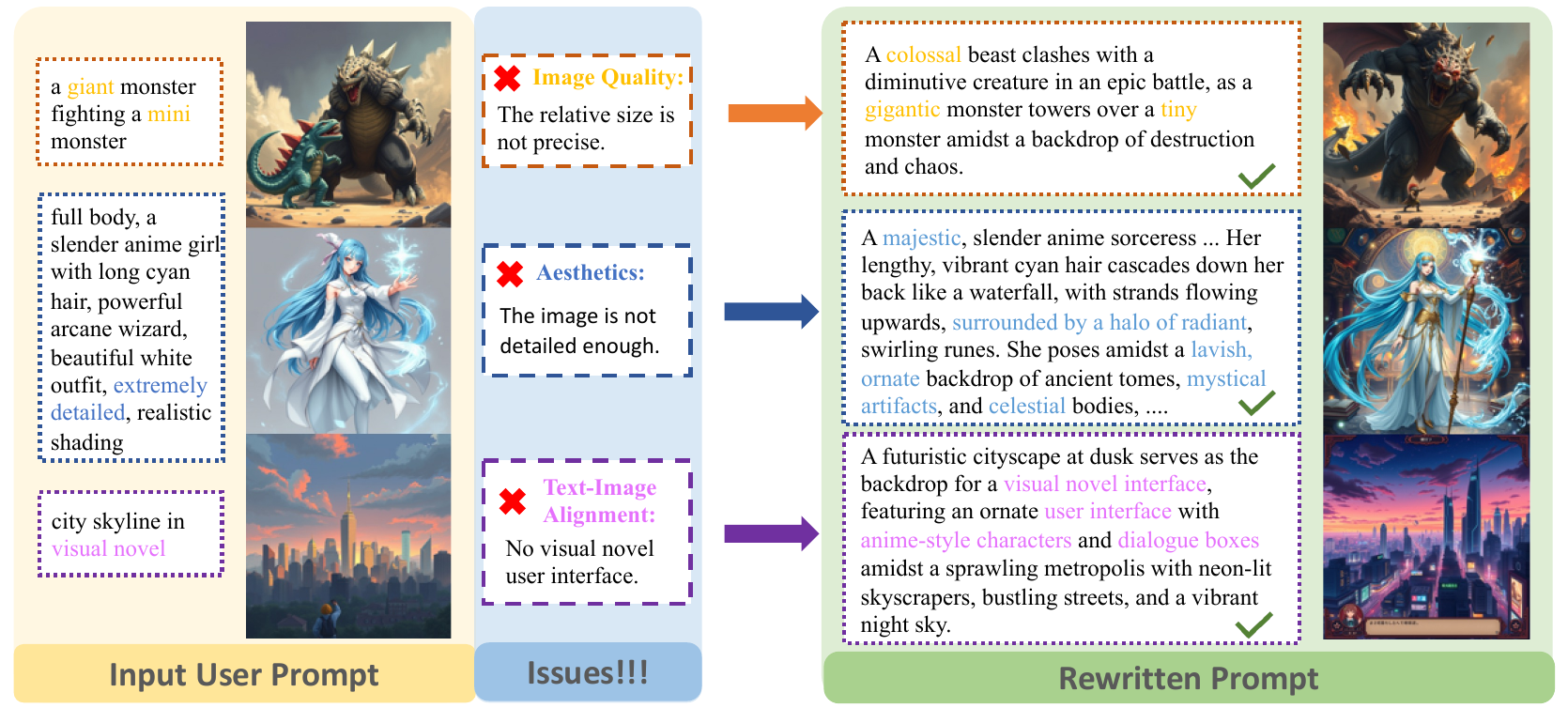}
    \caption{Overview of the \methodname\ framework for automated prompt rewriting. Input user prompts often lead to suboptimal generation results due to issues in image quality, aesthetics, or text-image alignment. \methodname\  rewrites these prompts to enhance detail and semantic clarity, resulting in significantly improved visual outputs across various categories.}
    \label{fig:placeholder}
    \vspace{-5mm}
\end{figure}
Text-to-image (T2I) generation has progressed rapidly along several fronts, including diffusion models~\cite{wu2025qwenimagetechnicalreport,flux2024,DBLP:conf/icml/EsserKBEMSLLSBP24}, autoregressive models~\cite{DBLP:journals/corr/abs-2501-17811,DBLP:journals/corr/abs-2409-18869}, and emerging paradigms such as next-scale prediction~\cite{DBLP:conf/nips/TianJYPW24}, all of which demonstrate strong performance in terms of image quality and fidelity. However, one important direction remains relatively underexplored: a simple, \emph{model-agnostic}, and \emph{training-free} pipeline that systematically addresses common shortcomings, such as weak text–image alignment and inconsistent aesthetics, by modifying only the input text.

In this work, we study an input-side strategy that treats the T2I model as a black box and refines the user prompt, leaving the model parameters unchanged. This direction carries a \emph{two-fold significance}: (i) from a practical perspective, real-world prompts are often short, vague, or underspecified. This creates a significant distributional gap between user inputs and the highly descriptive captions T2I models natively prefer. Rewriting these queries into clearer, more complete instructions mitigates alignment errors and stabilizes stylistic consistency; (ii) from a methodological perspective, it is of great interest to systematically study the achievable gain from optimizing the inputs alone.

Prior work refines prompts by using the In-Context Learning ability of LLMs to expand or restructure the input text~\cite{wu2025qwenimagetechnicalreport,hu2024ella}, and by employing human-in-the-loop systems~\cite{feng2023promptmagician,brade2023promptify} that iteratively adjust prompts with feedback from complex engineering pipelines and human evaluators. These methods are not fully automated and depend on substantial human effort. Other approaches~\cite{betker2023improving,DBLP:conf/acl/DattaKRA24} train dedicated rewriters with supervised fine-tuning (SFT) on curated pairs of short and refined prompts. However, because different T2I models encode different preferences for “good” prompts, collecting high-quality, model-specific annotations is costly, may not transfer well across backbones, and ties improvements to a particular target model and dataset.

To bridge this distributional gap without the prohibitive costs of SFT, we propose \methodname\ (Preference Optimization for Enhanced Text-to-Image generation), an automated, reinforcement learning-driven prompt rewriting framework. Rather than manually engineering prompts or retraining the T2I backbone, \methodname\ optimizes the prompt that feeds a \emph{frozen} T2I model. Concretely, we train a prompt rewriter using iterative Direct Preference Optimization (DPO)~\cite{DBLP:conf/nips/RafailovSMMEF23}. Starting from a short user-provided prompt, the rewriter generates a set of candidate refined prompts. Each candidate is then used to synthesize images with a frozen T2I model, which are subsequently evaluated by multimodal LLM judges according to a composite reward function that integrates (i) image quality, (ii) aesthetics, and (iii) text–image alignment. Pairwise preferences inferred from these reward scores are employed to update the rewriter iteratively via DPO. Empirical evaluations on real-world user prompts demonstrate that our approach attains state-of-the-art performance, while avoiding the substantial costs associated with SFT data collection and curation. We further summarize our contributions below:

\begin{itemize}
\item We introduce \methodname, an \textbf{RL-driven prompt rewriting framework} to enhance T2I model performance, which is both \textbf{model-agnostic} and \textbf{training-free} for the T2I generation models. Our policy optimizes only the input text, improving both diffusion and autoregressive T2I backbones \emph{without} SFT pairs, parameter updates, or architectural access.


\item Through extensive experiments across multiple LLMs, T2I models, and benchmarks, we show that \methodname\ consistently improves T2I performance in terms of image quality, aesthetics, and text–image alignment, achieving state-of-the-art results.

\item We conduct a systematic study of \methodname's transferability. Specifically, our RL-trained rewriter \emph{transfers} across diverse T2I backbones, quantifying cross-model robustness without per-model adaptation.
\end{itemize}

%% file: sec/2_related_work.tex
\vspace{-1mm}
\section{Related Work}
\vspace{-1mm}

Prompts play a pivotal role in determining the quality and fidelity of outputs in downstream tasks. Consequently, a growing body of research~\cite{DBLP:conf/acl/KongHZMB24, DBLP:conf/www/00120MKB24, DBLP:conf/aaai/WangZPG0025} has investigated techniques for prompt refinement in natural language processing. In the context of text-to-image (T2I) generation, prompt refinement methods can be broadly categorized as follows.

\vspace{-1mm}
\subsection{In-Context Learning}

In-Context Learning (ICL) has demonstrated strong effectiveness across a wide range of textual tasks. Extending this concept to text-to-image generation, Zeng et al.~\cite{zeng2024can} formally define T2I ICL and introduce a benchmark for evaluating the in-context capabilities of T2I models. Several recent models, including Image-Qwen~\cite{wu2025qwenimagetechnicalreport}, Emu2~\cite{sun2024generative}, and ELLA~\cite{hu2024ella}, report performance gains attributable to ICL. Furthermore, Lee et al.~\cite{DBLP:journals/access/LeeLBCLA24} apply in-context few-shot learning strategies to further enhance T2I generation quality. While effective, ICL relies heavily on prompt sensitivity and context window limits, motivating more systematic optimization approaches.

\vspace{-1mm}
\subsection{Chain-of-Thought Reasoning}

Chain-of-thought (CoT) reasoning has proven highly effective for complex text-based tasks. Guo et al.~\cite{DBLP:journals/corr/abs-2501-13926} provide the first systematic study on adapting CoT techniques to autoregressive image generation, introducing specialized reward models that iteratively verify and refine outputs during the generation process. Complementarily, ImageGen-CoT~\cite{DBLP:journals/corr/abs-2503-19312} proposes an automated pipeline for constructing high-quality datasets to fine-tune unified multimodal large language models, thereby enhancing their contextual reasoning abilities and improving image generation quality. Recent frameworks like UniGen~\cite{tian2025unigen} further explore bridging language and vision via unified generation pathways. In contrast to these methods, which often intervene directly in the multi-step generation or reasoning process, \methodname\ isolates the optimization entirely to the input prompt, treating the T2I backbone as a frozen environment.

\vspace{-1mm}
\subsection{Prompt Refinement}

\paragraph{Learning-free Methods.}
A number of works~\cite{feng2023promptmagician, brade2023promptify} explore interactive refinement approaches that leverage human input to produce more detailed and effective prompts for image synthesis. Manas et al. ~\cite{DBLP:journals/corr/abs-2403-17804} propose a backpropagation-free optimization strategy that automatically rewrites prompts to improve prompt–image consistency. Similarly, Chen et al.~\cite{chen2024tailored} present a method that exploits historical user interactions to rewrite prompts, thereby better aligning generated images with individual user preferences. However, these methods are not fully automated and often depend on substantial human effort or historical data.
\vspace{-1mm}

\paragraph{Supervised Fine-Tuning.}
A straightforward strategy for automated prompt refinement is supervised fine-tuning (SFT). For instance, DALL-E 3~\cite{betker2023improving} employs an auxiliary image captioner trained to generate high-quality captions, which are subsequently used to recaption images and enhance T2I training. Datta et al.~\cite{DBLP:conf/acl/DattaKRA24} introduce a Prompt Expansion framework that automatically generates multiple aesthetically rich prompt variants from a single user query, thereby improving generation quality. Despite their success, SFT approaches require the expensive curation of high-quality, model-specific annotations, which may not transfer well across different T2I backbones.

\vspace{-1mm}
\paragraph{Reinforcement Learning.}
Reinforcement learning (RL) has also been applied to prompt refinement. An early study by Hao et al.~\cite{DBLP:conf/nips/HaoC0W23} utilized PPO~\cite{DBLP:journals/corr/SchulmanWDRK17} to fine-tune a GPT-2 model for prompt rewriting. Building on this direction, Wu et al.~\cite{DBLP:conf/naacl/WuGW0W24} combined SFT and PPO to mitigate prompt toxicity and generate safer images. Lee et al.~\cite{lee2024parrot} proposed a multi-reward RL framework that jointly fine-tunes a T2I diffusion model and a prompt-expansion network. A concurrent work, RePrompt~\cite{wu2025reprompt}, explores RL-based refinement for T2I prompts. Our framework, \methodname, distinguishes itself by bypassing SFT entirely and employing Direct Preference Optimization (DPO) solely on the rewriter, avoiding the instability of PPO and the need to jointly update the T2I model. A detailed comparison between our method and RePrompt is provided in Appendix~\ref{appendix:diff}.

\vspace{-2mm}

%% file: sec/3_method_new.tex
\section{Method}
\label{sec:method}
\vspace{-1mm}

\begin{figure*}[h]
    \centering
    \includegraphics[width=0.99\linewidth]{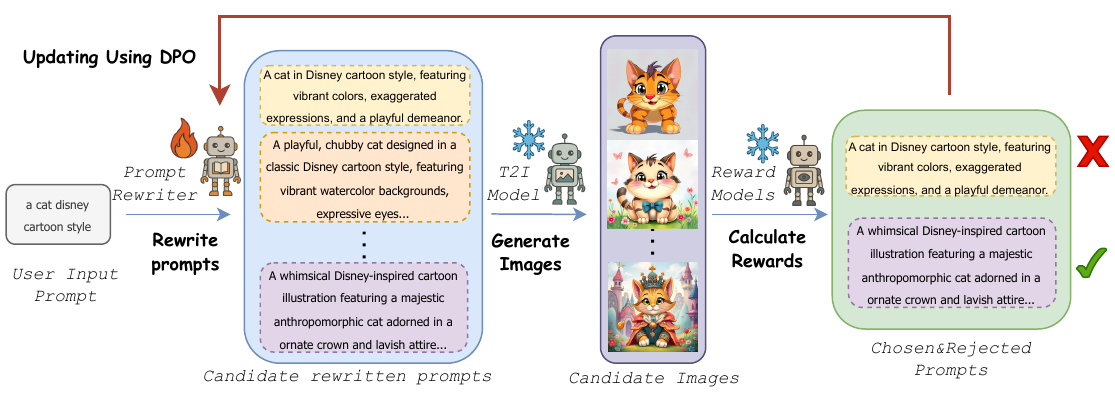}
    
    \caption{\methodname\ Training Pipeline. Given a user input prompt, the rewriter generates multiple candidate refinements, which are used by a frozen T2I model to synthesize images. Reward models evaluate these images to produce pairwise preferences. The chosen and rejected prompt pairs are then used to update the rewriter via Direct Preference Optimization, and the process is repeated iteratively for multiple rounds. The T2I model is kept frozen.}
    \vspace{-4mm}
    
    \label{fig:placeholder}
\end{figure*}

The core of \methodname\ is an iterative Direct Preference Optimization (DPO) pipeline that trains an LLM-based prompt rewriter without relying on Supervised Fine-Tuning (SFT). Our results demonstrate that the rewriter can successfully learn the complex task of prompt expansion and alignment purely through reinforcement learning. 

For each short input prompt, the policy generates $n$ candidate rewrites, which are then passed to a frozen T2I model to synthesize images. A multimodal LLM (MLLM) judge evaluates these images and provides pairwise preferences across multiple criteria, acting as the reward signal. The aggregated preferences yield a single winner--loser pair per prompt, which is used to optimize the DPO loss. This procedure is repeated over multiple rounds to progressively refine the rewriter's understanding of the T2I model's implicit preferences. The overall \methodname\ pipeline is illustrated in Figure~\ref{fig:placeholder}.

\subsection{Iterative DPO without SFT}

SFT typically requires model-specific data curation, which is both costly and prone to human biases tied to a particular backbone, thereby hindering transferability. Moreover, our preliminary experiments revealed that SFT diminishes the policy’s exploratory capacity, resulting in overfitting to the training prompts rather than exhibiting robust generalization at test time. To overcome these limitations, \methodname\ bypasses SFT entirely and trains the rewriter exclusively through iterative DPO.

Concretely, at each training round, the rewriter proposes multiple candidate rewrites for a user prompt $x$. Images synthesized from these rewrites by a frozen T2I model are scored to produce pairwise preferences. We then update the policy via DPO using the following objective:

\begin{equation}
\mathcal{L}_\text{DPO}(\pi_{\theta}; \pi_{\text{ref}})
= -\mathbb{E}_{(x, y_w, y_l)\sim \mathcal{D}}
\Biggl[
\log \sigma\Bigl(
\beta \log \frac{\pi_{\theta}(y_w\mid x)}{\pi_{\text{ref}}(y_w\mid x)}
- \beta \log \frac{\pi_{\theta}(y_l\mid x)}{\pi_{\text{ref}}(y_l\mid x)}
\Bigr)
\Biggr] \label{eq:dpo}
\end{equation}
where $x$ is the input prompt, $y_w$ and $y_l$ represent the chosen and rejected rewritten prompts respectively, and $\pi_{\text{ref}}$ is the reference model distribution.

We intentionally do not adopt GRPO~\cite{shao2024deepseekmath,guo2025deepseek} or its variants, as these methods are most effective when a reliable scalar reward is available. Instead, we rely on \emph{pairwise} image comparisons to significantly reduce variance, which align natively with DPO. Furthermore, Tong et al.~\cite{tong2025delving} report that in image generation settings involving reasoning, DPO yields stronger in-domain performance than GRPO, empirically validating our choice for the \methodname\ framework.

\subsection{Reward Design}

We adopt the \emph{MLLM-as-a-judge} paradigm and employ Qwen2.5-VL-72B-Instruct as the evaluation model. During training, the judge is provided with an instruction alongside \emph{two} candidate images, outputting a pairwise preference or a tie. 

To comprehensively capture the primary aspects of user intent and visual fidelity, we define four distinct reward dimensions: (1) image quality, $r_{\text{Quality}}$; (2) general image–text alignment, $r_{\text{General-Alignment}}$; (3) physical image–text alignment, $r_{\text{Physical-Alignment}}$; and (4) aesthetics, $r_{\text{Aesthetics}}$.

\paragraph{Image Quality.}  
The image quality reward evaluates whether the generated image is visually plausible across five dimensions: (1) correctness of human or animal body parts, (2) plausibility of geometric details (e.g., absence of unintended wavy lines, consistency in symmetrical objects), (3) adherence to basic physical laws such as reflections and shadows, (4) semantic and logical coherence of the scene, and (5) correctness of fine-grained details without noticeable noise or distortions.

\paragraph{General Image–Text Alignment.}  
This reward evaluates the degree to which an image adheres to the input prompt at a holistic level. Because assessing a complex prompt directly can be challenging for the judge, we employ a two-step procedure:
\begin{enumerate}
    \item \emph{Decomposition:} Given the input prompt, the judge generates a concise list of key questions capturing its essential components. For instance, for the prompt “four apples on a table,” the decomposition yields: “Are there four apples?”, “Is there a table?”, and “Are the four apples on the table rather than below it?”
    \item \emph{Comparison:} The decomposed questions, alongside the candidate image pair, are provided to the judge to produce a pairwise preference decision.
\end{enumerate}

\paragraph{Physical Image–Text Alignment.}
This reward targets concrete physical relationships that text-to-image models frequently fail to capture. It evaluates: (i) the correctness of object counts, (ii) spatial relations (e.g., left/right, on/under, in front/behind), and (iii) attribute binding (i.e., whether the specified color, size, or style is accurately associated with the correct object). 

\paragraph{Image Aesthetics.}
The aesthetics reward assesses the relative visual appeal and stylistic quality of two candidate images, taking into account composition, richness of detail, and overall artistic merit.

\vspace{1em}
Based on these rewards, we train two distinct variants of \methodname: a \emph{general rewriter}, which prioritizes semantic faithfulness and overall image quality, and an \emph{aesthetics rewriter}, which emphasizes visual appeal while maintaining baseline alignment and quality. We train these two policies by combining the rewards as follows:


\vspace{-2mm}

\begin{equation}
\left\{
\begin{aligned}
r^{\text{General}} &= r_{\text{Quality}} + r_{\text{General-Alignment}} + r_{\text{Physical-Alignment}} \\
r^{\text{Aesthetics}} &= r_{\text{Quality}} + r_{\text{General-Alignment}} + r_{\text{Physical-Alignment}} + r_{\text{Aesthetics}}
\end{aligned}
\right.
\end{equation}
More details on the specific prompts designed for the MLLM-as-a-judge reward are provided in Appendix~\ref{appendix:reward}.
\vspace{-1mm}

\subsection{Selection of Chosen and Rejected Prompts}
\vspace{-2mm}
\begin{algorithm}[h]               
  \caption{Selection of chosen and rejected rewritten prompts}      
  \label{alg:get_chosen_and_rejected}              
  \begin{algorithmic}       
    \Require Original prompt $x$, rewritten candidates $\{y_1,y_2,\dots,y_n\}$, text-to-image model $g$, reward models $\{\text{RM}_k\}_{k=1}^K$
    \For{$i = 1$ to $n$}
        \State Generate image $\text{Img}_i \gets g(y_i)$
        \State Initialize rewards $r_i^k \gets 0$ for all $k = 1,\dots,K$
    \EndFor
    \For{$i = 1$ to $n$}
        \For{$j = 1$ to $n$ \textbf{such that} $j \neq i$}
            \For{$k = 1$ to $K$}
                \If{$\text{Img}_i \succ \text{Img}_j \mid \text{RM}_k, x$}
                    \State $r_i^k \gets r_i^k + 1$, \quad $r_j^k \gets r_j^k - 1$
                \ElsIf{$\text{Img}_i \prec \text{Img}_j \mid \text{RM}_k, x$}
                    \State $r_i^k \gets r_i^k - 1$, \quad $r_j^k \gets r_j^k + 1$
                \EndIf
                \Comment{No update if $\text{Img}_i \approx \text{Img}_j \mid \text{RM}_k, x$}
            \EndFor
        \EndFor
        \State Aggregate rewards: $R_i \gets \sum_{k=1}^K r_i^k$
    \EndFor
    \State Select chosen prompt $y_w \gets y_{i_w}$, $i_w \gets \arg\max_{i} R_i$
    \State Select rejected prompt $y_l \gets y_{i_l}$, $i_l \gets \arg\min_{i} R_i$
    \State \Return $(y_w, y_l)$
  \end{algorithmic}
\end{algorithm}
\vspace{-1mm}

\begin{algorithm}[h]    
  \caption{Iterative DPO for \methodname}      
  \label{alg:iterative_dpo}              
  \begin{algorithmic}          
    \Require Base LLM $f_\theta$, training prompt set $\mathcal{D}$,  samples-per-round $m$, candidates-per-prompt $n$
    \For{each round}
    \State Update reference policy $f_{ref}=f_\theta$.
    \State Sample prompts for each round: $\{x_i\}_{i=1}^m \sim \mathcal{D}$.
        \For{$i=1$ to $m$}
        \State Generate rewritten prompts: $\{y_i^j\}_{j=1}^{n} \sim f_{\theta}(\,\cdot \mid x_i)$.
        \State Get chosen and rejected rewritten prompts $(y_i^w,y_i^l)$ with Alg.~\ref{alg:get_chosen_and_rejected}.
        \EndFor
        \For{each step}
        \State Sample a batch $B$ from $\{(x_i,y_i^w,y_i^l)\}_{i=1}^{m}$,
        \State Update $f_\theta$ with $f_\text{ref}$ and $B$ using Eq.~\ref{eq:dpo}.
        \EndFor
    \EndFor
    \State \Return $f_\theta$
  \end{algorithmic}
\end{algorithm}
\vspace{-1mm}
For each original prompt, we construct a single chosen-rejected pair for DPO training using pairwise image comparisons, which provide much more reliable supervision than pointwise scoring.  

Given a prompt $x$, \methodname\ samples $n$ rewritten candidates and renders $n$ corresponding images. From these, we form all $n(n-1)$ \emph{ordered} pairs. Each ordered pair is evaluated by all reward models associated with the current rewriter variant. For each comparison, the reward model issues a preference: the preferred image receives $+1$, the other receives $-1$, and a tie assigns $0$ to both.  

We aggregate these votes across all reward models and all ordered pairs to obtain a cumulative score for each rewritten candidate. The candidate with the highest score is selected as the chosen prompt $y_w$, while the one with the lowest score is designated as the rejected prompt $y_l$. The triplet $(x, y_w, y_l)$ is subsequently used as a DPO training instance. The complete procedure for selecting the chosen and rejected prompts is summarized in Algorithm~\ref{alg:get_chosen_and_rejected}.

\subsection{Iterative DPO Pipeline}

At each iteration of the DPO training, we draw a batch of user prompts and employ the current rewriter policy to generate $n$ candidate rewrites for each prompt. Subsequently, Algorithm~\ref{alg:get_chosen_and_rejected} is applied to the pairwise image comparisons to identify a single preferred and a single rejected rewrite per prompt. The policy is then updated by minimizing the DPO loss defined in Equation~\ref{eq:dpo}. The overall iterative training process for \methodname\ is summarized in Algorithm~\ref{alg:iterative_dpo}.


%% file: sec/4_exp.tex
\section{Experiments}
\vspace{-2mm}
To evaluate the effectiveness of \methodname, we conduct comprehensive experiments across multiple text-to-image architectures and benchmarks. Detailed configurations, including the specific LLMs used for the prompt rewriter, the frozen T2I models, training datasets, baseline comparisons, and hyperparameters, are provided in Appendix~\ref{appendix:setting}.

\subsection{Benchmark Results} 

\begin{table}[htbp]
\centering

\caption{GenEval results compared with state-of-the-art methods. Using the general rewriter with Llama-3-70B-Instruct as the LLM backbone, \methodname\ achieves the best overall GenEval score (higher is better). \textbf{Bold} is the best and \underline{underline} the second-best.  $^\dag$Results reported by Wu et al.~\cite{wu2025reprompt}. $^\ddag$Results reported by Guo et al.~\cite{DBLP:journals/corr/abs-2501-13926}.}

\resizebox{\textwidth}{!}{%
\setlength{\tabcolsep}{2.5pt}
\begin{tabular}{@{}l|l|cccccc|c@{}}
\toprule
Method & \begin{tabular}[c]{@{}l@{}}T2I\\ Backbone\end{tabular} & \begin{tabular}[c]{@{}c@{}}Single\\ object\end{tabular} & \begin{tabular}[c]{@{}c@{}}Two\\ object\end{tabular} & Counting & Colors & Position & \begin{tabular}[c]{@{}c@{}}Color\\ attribution\end{tabular} & Overall \\ \midrule
Baseline$^\ddag$ & Show-o & 0.95 & 0.52 & 0.49 & 0.82 & 0.11 & 0.28 & 0.53 \\
Baseline & FLUX.1-dev & 1.00 & 0.87 & 0.76 & 0.84 & 0.22 & 0.49 & 0.70 \\ \midrule
PARM$^\ddag$ & Show-o & \underline{0.99} & 0.77 & 0.68 & 0.86 & 0.29 & 0.45 & 0.67 \\
PARM++$^\ddag$ & Show-o & \underline{0.99} & 0.71 & 0.69 & \textbf{0.95} & 0.36 & \underline{0.49} & 0.70 \\
Idea2Img$^\dag$ & FLUX.1-dev & - & - & - & - & - & - & 0.69 \\
Zero-Shot & FLUX.1-dev & 0.98 & \underline{0.89} & 0.71 & 0.82 & 0.47 & \underline{0.49} & 0.73 \\
ICL & FLUX.1-dev & \textbf{1.00} & \underline{0.89} & \textbf{0.80} & 0.83 & 0.54 & 0.41 & 0.74 \\
RePrompt$^\dag$ & FLUX.1-dev & 0.98 & 0.87 & 0.77 & 0.85 & \textbf{0.62} & 0.49 & \underline{0.76} \\
\rowcolor{blue!10} \methodname\ (Ours) & FLUX.1-dev & \textbf{1.00} & \textbf{0.95} & \underline{0.78} & \underline{0.88} & \underline{0.58} & \textbf{0.56} & \textbf{0.79} \\ \bottomrule
\end{tabular}%
}
\label{tab:geneval_baseline}
\end{table}

\begin{table}[t]
\centering

\caption{GenEval results for the general rewriter across T2I backbones. \methodname\ consistently outperforms the original prompts and ICL baselines, and performance improves with LLM size; Llama-3-70B-Instruct achieves the best results.}
\vspace{-1mm}
\label{tab:geneval_ours}
\begin{tabular}{@{}l|cccccc|c@{}}
\toprule
Model & \begin{tabular}[c]{@{}c@{}}Single\\ object\end{tabular} & \begin{tabular}[c]{@{}c@{}}Two\\ object\end{tabular} & Counting & Colors & Position & \begin{tabular}[c]{@{}c@{}}Color\\ attribution\end{tabular} & Overall \\ \midrule
FLUX.1-schnell & 0.99 & 0.87 & 0.61 & 0.79 & 0.35 & 0.43 & 0.67 \\
+ ICL & 0.99 & 0.87 & 0.55 & 0.81 & 0.50 & 0.50 & 0.69 \\
+ \methodname\ (Llama 3 3B) & \textbf{1.00} & 0.89 & 0.63 & 0.81 & 0.47 & \textbf{0.59} & 0.73 \\
+ \methodname\ (Llama 3 8B) & \textbf{1.00} & \textbf{0.93} & 0.58 & \textbf{0.86} & 0.55 & 0.55 & 0.74 \\
\rowcolor{blue!10} + \methodname\ (Llama 3 70B) & 0.99 & 0.91 & \textbf{0.65} & 0.77 & \textbf{0.64} & 0.57 & \textbf{0.75} \\ \midrule
FLUX.1-dev & \textbf{1.00} & 0.87 & 0.76 & 0.84 & 0.22 & 0.49 & 0.70 \\
+ ICL & \textbf{1.00} & 0.89 & \textbf{0.80} & 0.83 & 0.54 & 0.41 & 0.74 \\
\rowcolor{blue!10} + \methodname\ (Llama 3 70B) & \textbf{1.00} & \textbf{0.95} & 0.78 & \textbf{0.88} & \textbf{0.58} & \textbf{0.56} & \textbf{0.79} \\ \midrule
SD-3.5-medium & \textbf{1.00} & 0.84 & \textbf{0.68} & 0.85 & 0.24 & \textbf{0.61} & 0.70 \\
+ ICL & 0.96 & \textbf{0.92} & 0.60 & \textbf{0.86} & 0.42 & 0.54 & 0.72 \\
\rowcolor{blue!10} + \methodname\ (Llama 3 70B) & \textbf{1.00} & \textbf{0.92} & \textbf{0.68} & 0.85 & \textbf{0.57} & 0.56 & \textbf{0.76} \\ \bottomrule
\end{tabular}
\vspace{-1mm}

\end{table}

\begin{table}[htbp]
\centering

\caption{Results on T2I-CompBench++, TIFA-Benchmark, and FID on MS COCO 30K using the general rewriter with Llama-3-70B-Instruct as the LLM backbone. \methodname\ improves text–image alignment and image quality across all evaluated T2I backbones.}
\vspace{-1mm}
\label{tab:benchmarks}
\resizebox{\linewidth}{!}{%
\begin{tabular}{@{}l|cccccc|c|c@{}}
\toprule
 & \multicolumn{6}{c|}{T2I-CompBench++} & TIFA & COCO \\ \cmidrule(l){2-9} 
 & Color$\uparrow$ & Shape$\uparrow$ & Texture$\uparrow$ & Spatial$\uparrow$ & Numeracy$\uparrow$ & Complex$\uparrow$ & Score$\uparrow$ & FID$\downarrow$ \\ \midrule
FLUX.1-schnell & 0.7492 & 0.5673 & 0.6911 & 0.2754 & 0.6062 & 0.3614 & 0.8803 & 20.57 \\
\rowcolor{blue!10} + \methodname & \textbf{0.7614} & \textbf{0.6010} & \textbf{0.7063} & \textbf{0.3216} & \textbf{0.6161} & \textbf{0.3713} & \textbf{0.8868} & \textbf{17.76} \\ \midrule
FLUX.1-dev & 0.7647 & 0.5053 & 0.6336 & 0.2763 & 0.6130 & 0.3586 & 0.8572 & 24.38 \\
\rowcolor{blue!10} + \methodname & \textbf{0.7978} & \textbf{0.5834} & \textbf{0.6929} & \textbf{0.3206} & \textbf{0.6343} & \textbf{0.3694} & \textbf{0.8809} & \textbf{19.57} \\ \midrule
SD-3.5-medium & 0.7988 & 0.5800 & 0.7206 & 0.2889 & 0.6033 & 0.3614 & 0.8782 & 17.81 \\
\rowcolor{blue!10} + \methodname & \textbf{0.8040} & \textbf{0.5855} & \textbf{0.7346} & \textbf{0.3322} & \textbf{0.6320} & \textbf{0.3707} & \textbf{0.8878} & \textbf{17.13} \\ \midrule
JanusPro & 0.5294 & 0.3247 & 0.4168 & 0.1579 & 0.4380 & 0.3778 & 0.8457 & 19.28 \\
\rowcolor{blue!10} + \methodname & \textbf{0.7861} & \textbf{0.5892} & \textbf{0.7220} & \textbf{0.2773} & \textbf{0.5983} & \textbf{0.3911} & \textbf{0.8845} & \textbf{16.71} \\ \bottomrule
\end{tabular}%
}

\end{table}

We benchmark \methodname\ against recent state-of-the-art methods on GenEval. Under a standardized evaluation protocol (Table~\ref{tab:geneval_baseline}), \methodname\ achieves the highest overall GenEval score, outperforming all competing methods across both model families by a substantial margin.

We further test \methodname\ on more datasets with different T2I backbones.
As shown in Tables~\ref{tab:geneval_ours} and~\ref{tab:benchmarks}, \methodname\ achieves strong performance on GenEval, T2I-CompBench++, and the TIFA-Benchmark in terms of text–image alignment, and attains lower FID scores on MS-COCO-30K for image quality. Across all benchmarks, our approach consistently improves alignment, for example increasing the FLUX.1-dev  GenEval score from 0.70 to 0.79, while simultaneously enhancing image quality as reflected by reduced FID.

\subsection{MLLM-as-a-Judge Evaluations}

\begin{table}[htbp]
\centering

\caption{Pick-a-Pic v2 results with Llama-3-70B-Instruct as the LLM backbone. Reported values are GPT-4o–judged win rates vs. DALL·E 3. The general rewriter achieves the highest text–image alignment while also improving quality and aesthetics; the aesthetics rewriter delivers the best aesthetics while maintaining quality and alignment. \textbf{Bold} is the best and \underline{underline} the second-best.}
\vspace{-1mm}
\label{tab:main_pick-a-pic}
\begin{tabular}{@{}l|cccc|c@{}}
\toprule
 & \begin{tabular}[c]{@{}c@{}}Image\\ Quality\end{tabular} & \begin{tabular}[c]{@{}c@{}}Image\\ Aesthetics\end{tabular} & \begin{tabular}[c]{@{}c@{}}Image-Text\\ Alignment\end{tabular} & Average & LAION \\ \midrule
Dalle-3 & 0.500 & 0.500 & 0.500 & 0.500 & 6.19 \\ \midrule
FLUX.1-schnell & 0.469 & 0.314 & 0.419 & 0.401 & 6.07 \\
+ ICL & 0.475 & 0.307 & 0.422 & 0.401 & 6.07 \\
+ \methodname\ (General Rewriter) & \underline{0.494} & \underline{0.476} & \textbf{0.561} & \underline{0.510} & \underline{6.08} \\
+ \methodname\ (Aesthetics Rewriter) & \textbf{0.495} & \textbf{0.818} & \underline{0.424} & \textbf{0.579} & \textbf{6.39} \\ \midrule
FLUX.1-dev & 0.513 & 0.350 & 0.360 & 0.408 & \underline{6.22} \\
+ ICL & 0.531 & 0.363 & \underline{0.457} & 0.450 & 6.17 \\
+ \methodname\ (General Rewriter) & \underline{0.536} & \underline{0.491} & \textbf{0.575} & \underline{0.534} & 6.06 \\
+ \methodname (Aesthetics Rewriter) & \textbf{0.576} & \textbf{0.800} & 0.391 & \textbf{0.589} & \textbf{6.41} \\ \midrule
SD-3.5-medium & 0.424 & 0.230 & 0.425 & 0.359 & 5.74 \\
+ ICL & \underline{0.487} & 0.327 & \underline{0.455} & 0.423 & 5.92 \\
+ \methodname (General Rewriter) & \textbf{0.504} & \underline{0.456} & \textbf{0.542} & \underline{0.501} & \underline{5.95} \\
+ \methodname (Aesthetics Rewriter) & 0.481 & \textbf{0.799} & 0.401 & \textbf{0.560} & \textbf{6.23} \\ \midrule
JanusPro & 0.193 & 0.181 & \underline{0.392} & 0.255 & 5.89 \\
+ ICL & \textbf{0.213} & 0.180 & 0.335 & 0.243 & \underline{5.97} \\
+ \methodname (General Rewriter) & 0.183 & \underline{0.268} & \textbf{0.421} & \underline{0.291} & 5.95 \\
+ \methodname (Aesthetics Rewriter) & \underline{0.200} & \textbf{0.612} & 0.317 & \textbf{0.377} & \textbf{6.23} \\ \bottomrule
\end{tabular}
\end{table}

\begin{table*}[htbp]
\centering
\caption{PartiPrompts~\cite{DBLP:journals/tmlr/YuXKLBWVKYAHHPLZBW22} results. Values are GPT-4o–judged win rates (higher is better) of our rewritten prompts vs. the original prompts. \methodname\ consistently improves image quality, alignment, and aesthetics.}
\label{tab:partiprompts}
\begin{tabular}{@{}l|cccc@{}}
\toprule
 & \begin{tabular}[c]{@{}c@{}}Image\\ Quality\end{tabular} & \begin{tabular}[c]{@{}c@{}}Image\\ Aesthetics\end{tabular} & \begin{tabular}[c]{@{}c@{}}Image-Text\\ Alignment\end{tabular} & Average \\ \midrule
FLUX.1-schnell & 0.524 & 0.583 & 0.592 & 0.566 \\
FLUX.1-dev & 0.510 & 0.584 & 0.625 & 0.573 \\
SD-3.5-medium & 0.599 & 0.743 & 0.542 & 0.628 \\
JanusPro & 0.616 & 0.674 & 0.687 & 0.659 \\ \bottomrule
\end{tabular}
\end{table*}

On the Pick-a-Pic v2 dataset, using GPT-4o as a pairwise judge, \methodname\ consistently outperforms the original test prompts. Crucially, these improvements are achieved entirely on the input side, requiring no additional training or parameter updates to the underlying T2I backbones. As reported in Table~\ref{tab:main_pick-a-pic}, the \emph{general rewriter} variant attains the highest win rate for text-image alignment while simultaneously enhancing overall image quality and aesthetics. Conversely, the \emph{aesthetics rewriter} is highly effective at maximizing visual appeal, achieving a striking aesthetics win rate of 0.818, while still maintaining competitive performance on both image quality and alignment.

These robust gains extend to the PartiPrompts benchmark (detailed in Table~\ref{tab:partiprompts}). GPT-4o evaluations demonstrate that \methodname\ delivers consistent improvements across all tested T2I backbones. Specifically, the win rates of our rewritten prompts compared to the original, underspecified user inputs exceed 0.5 across every evaluation dimension.

We further observe that simple in-context learning (ICL) prompts frequently improve generation quality. Contemporary T2I models often underperform when conditioned on short or underspecified prompts, which are prevalent in real-world usage, whereas straightforward prompt expansions provide consistent benefits. This phenomenon likely arises from a distributional gap, as training captions are generally longer and more descriptive than prompts provided by real users.


\vspace{-2mm}

\subsection{Human Evaluations}

\vspace{-2mm}
To verify that our automated MLLM-as-a-judge metrics accurately reflect true human preference, we conducted a rigorous human evaluation study. As shown in Table~\ref{tab:human_evaluation}, human annotators consistently prefer images generated using \methodname\ refined prompts over the original baselines across all three dimensions. Notably, the human judgments closely mirror the GPT-4o evaluations. This strong correlation not only confirms the real-world effectiveness of our framework, but also validates the evaluation process.

\begin{table}[htbp]
\centering
\caption{Human evaluation and MLLM-as-a-judge results using the FLUX.1-dev backbone with Llama-3-70B-Instruct on the Pick-a-Pic v2 dataset. Values represent the pairwise win rates of \methodname\ refined prompts against the original user prompts. The human annotator preferences closely align with the GPT-4o automated judge, demonstrating consistent improvements in image quality, aesthetics, and text-image alignment.}
\label{tab:human_evaluation}
\begin{tabular}{@{}l|l|ccc|c@{}}
\toprule
 & Evaluator & \begin{tabular}[c]{@{}c@{}}Image\\ Quality\end{tabular} & \begin{tabular}[c]{@{}c@{}}Image\\ Aesthetics\end{tabular} & \begin{tabular}[c]{@{}c@{}}Image-Text\\ Alignment\end{tabular} & Average \\ \midrule
FLUX.1-dev & \multicolumn{1}{c|}{-} & 0.500 & 0.500 & 0.500 & 0.500 \\ \midrule
+ \methodname & GPT-4o & 0.523 & 0.561 & 0.658 & 0.581 \\
+ \methodname & Human & 0.552 & 0.561 & 0.618 & 0.577 \\ \bottomrule
\end{tabular}
\end{table}

\subsection{Qualitative Analysis}
\vspace{-2mm}
We provide qualitative examples in Appendix~\ref{appendix:case_study} that trace the evolution of rewritten prompts throughout the iterative DPO training process. Additionally, we visualize the stylistic differences between the outputs of the general rewriter, which prioritizes faithfulness, and the aesthetics rewriter, which focuses on visual appeal. Detailed examples can be found in Appendix Table~\ref{tab:aes_case_study}.

%% file: sec/5_ablation.tex
\section{Analyses}

\subsection{Reward Type}


Table~\ref{tab:ablation} presents an ablation study of the reward terms. Removing the quality reward substantially decreases the image quality win rate, from 0.494 to 0.364. Excluding either the general or physical alignment rewards reduces alignment, from 0.561 to 0.521 and to 0.504 respectively, confirming that each reward effectively drives its intended objective. Incorporating the aesthetics reward yields a pronounced improvement in aesthetics, increasing from 0.476 to 0.818, but simultaneously reduces alignment from 0.561 to 0.424, highlighting a trade-off. Qualitative analysis in Table~\ref{tab:aes_case_study} suggests that the aesthetics reward often enriches scenes by introducing additional objects or ornamentation, which can dilute the main subject and thereby weaken alignment. To accommodate different objectives, we report two variants: a general rewriter optimized for faithfulness and overall quality, and an aesthetics rewriter tailored to visual appeal.

\subsection{Rewritten Prompt Length}

Figure~\ref{fig:rewritten_length} illustrates a steady increase in the average token length of rewritten prompts over training in \methodname. The rewriter progressively incorporates additional specificity, such as attributes, relations, and constraints, and this increase in length correlates with higher win rates.


\subsection{Full Finetune vs. LoRA}

\begin{figure*}
    \centering
    \includegraphics[width=0.98\linewidth]{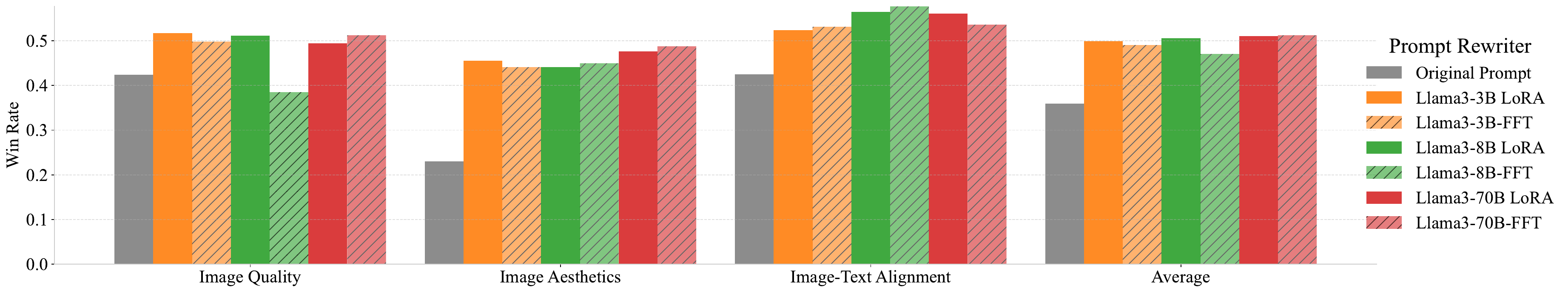}
    \vspace{-1mm}
    \caption{Full finetune vs. LoRA. win rates against DALL-E 3. Evaluated on the Pick-a-pic v2 datasets with GPT-4o judge. We use Llama3-70B-Instruct as the general rewriter.}
    \vspace{-3mm}
    \label{fig:fft_vs_lora}
\end{figure*}


We conduct a controlled comparison between LoRA and full-model finetuning under identical experimental conditions, as shown in Figure~\ref{fig:fft_vs_lora}. Full-model finetuning does not yield consistent improvements, whereas LoRA attains comparable performance while requiring substantially less memory and computation. These results suggest that LoRA constitutes a strong and practical choice for the training in \methodname.

%% file: sec/6_analyses.tex
\vspace{-1mm}
\subsection{LLM Backbones}
\label{sec:llm_backbones}
\vspace{-1mm}
To understand the impact of the rewriter's architectural family and reasoning capacity, we evaluated \methodname\ using three distinct LLM backbones. As detailed in Table~\ref{tab:llm_backbones}, all \methodname\ variants consistently surpass the FLUX.1-schnell baseline across every metric. 

Llama-3-70B-Instruct achieves the highest overall average. The comparatively lower aesthetic performance of the DeepSeek model suggests that heavy reasoning-oriented tuning may occasionally conflict with the exploratory prompt enrichment required to maximize visual appeal. These results confirm that while \methodname\ is effective across various LLMs, the choice of backbone allows practitioners to trade off between text-image alignment and aesthetic enhancement.

\begin{table}[h]
\centering
\caption{Performance comparison of \methodname\ across different LLM backbones using the FLUX.1-schnell T2I model. Values represent GPT-4o-judged pairwise win rates against DALL·E 3. While Llama 3 70B achieves the best overall text--image alignment and average score, Qwen 3 32B demonstrates highly efficient performance, securing the highest image quality score despite a smaller parameter count. \textbf{Bold} indicates the best score in each column.}
\vspace{-1mm}
\label{tab:llm_backbones}
\begin{tabular}{@{}l|ccc|c@{}}
\toprule
 & \begin{tabular}[c]{@{}c@{}}Image\\ Quality\end{tabular} & \begin{tabular}[c]{@{}c@{}}Image\\ Aesthetics\end{tabular} & \begin{tabular}[c]{@{}c@{}}Image--Text\\ Alignment\end{tabular} & Average \\ \midrule
FLUX.1-schnell & 0.469 & 0.314 & 0.419 & 0.401 \\ \midrule
+ \methodname\ (Qwen 3 32B) & \textbf{0.517} & 0.424 & 0.534 & 0.492 \\
+ \methodname\ (DeepSeek 70B) & 0.502 & 0.380 & 0.506 & 0.463 \\
+ \methodname\ (Llama 3 70B) & 0.494 & \textbf{0.476} & \textbf{0.561} & \textbf{0.510} \\ \bottomrule
\end{tabular}
\end{table}

\begin{figure}[]
    \centering
    \begin{subfigure}[b]{0.99\textwidth}
        \centering
        \includegraphics[width=\linewidth]{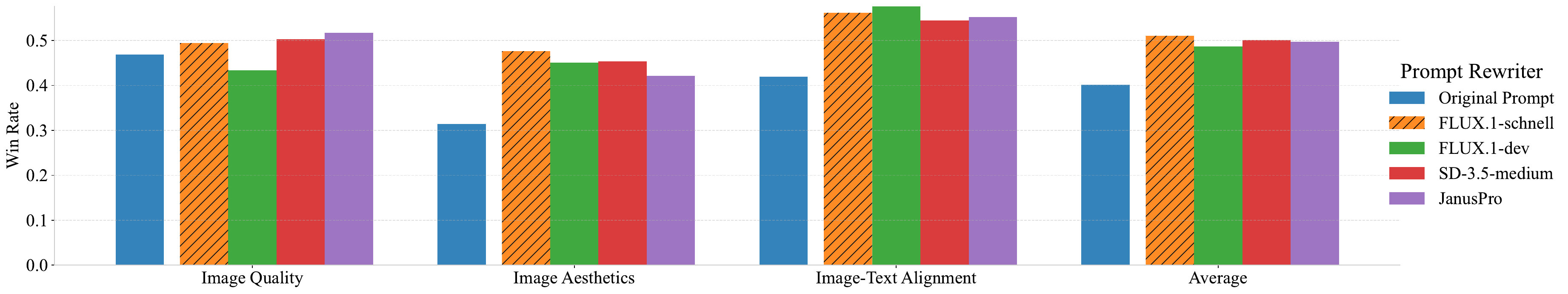}
        \caption{Win rate against DALL-E 3 using FLUX.1-schnell as test-time T2I.}
    \end{subfigure}
    \begin{subfigure}[b]{0.99\textwidth}
        \centering
        \includegraphics[width=\linewidth]{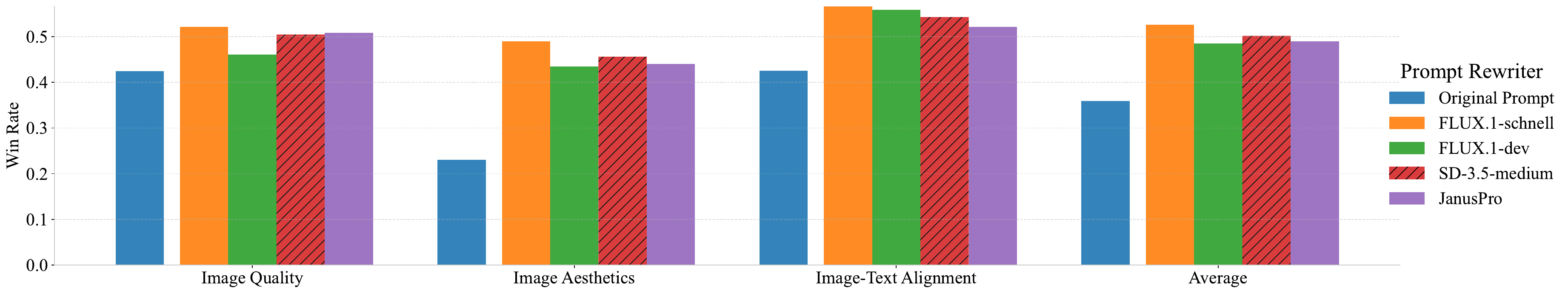}
        \caption{Win rate against DALL-E 3 using SD-3.5-medium as test-time T2I.}
    \end{subfigure}
    \caption{Win rates for rewriters trained on various T2I backbones but evaluated on a fixed target, either (a) FLUX.1-schnell or (b) SD-3.5-medium. Evaluation is on the Pick-a-pic v2 dataset using a GPT-4o judge. Results show strong transferability.}
    \vspace{-2mm}
    \label{fig:transfer}
\end{figure}



\vspace{-1mm}
\subsection{Transferability}


We evaluate transferability by training multiple rewriters, each paired with a different T2I backbone, and subsequently freezing the rewriter for testing on a fixed target backbone. We consider two target models, FLUX.1-schnell and Stable Diffusion 3.5. For each target, we report performance in three conditions: when the training and target T2I backbones are identical, when they differ, and when using the original unmodified prompt as a baseline. Results are presented in Figure~\ref{fig:transfer} and Table~\ref{tab:geneval_transfer}.

Across both targets, rewritten prompts consistently improve over the original prompts, irrespective of whether the training and testing backbones match. Performance differences under train–test mismatch are generally small and often comparable to the matched setting. These findings suggest that the rewriter acquires prompt refinements that generalize across T2I backbones, enabling reuse without per-model retraining. We hypothesize that this transferability arises because T2I models are commonly trained on image–caption pairs from similar distributions, rendering our rewrites broadly portable.

%% file: sec/7_conclusion.tex
\vspace{-1mm}
\section{Conclusion}
\vspace{-1mm}
In this work, we introduce \methodname, an automated, RL-driven prompt rewriting framework designed to bridge the distributional gap between underspecified user prompts and the highly descriptive captions preferred by text-to-image models. By leveraging iterative Direct Preference Optimization guided by a composite MLLM-as-a-judge reward system, our approach learns to optimize inputs without relying on costly supervised fine-tuning. Consequently, \methodname\ consistently improves image quality, text--mage alignment, and aesthetics across diverse T2I backbones while leaving their underlying parameters entirely frozen. Extensive experiments further demonstrate the framework's strong cross-model transferability. \methodname\ establishes input-side optimization as a highly practical, model-agnostic strategy for advancing T2I systems.

%% file: sec/X_suppl.tex
\clearpage
\setcounter{page}{1}

\section{Experiment Setting Details}
\label{appendix:setting}

\subsection{Models and Datasets}


We conduct training on two datasets: Pick-a-Pic v2~\cite{kirstain2023pick}, which consists of real user prompts, and T2I-CompBench++~\cite{huang2025t2i}, which provides compositional prompts designed to evaluate text–image alignment. For T2I-CompBench++, we adopt the official training split. Since Pick-a-Pic v2 does not provide predefined splits, we partition the data ourselves and utilize only the training portion. In total, our training corpus contains approximately 60k prompts. For the Pick-a-Pic v2 test split, we keep 500 prompts for evaluations. As our approach does not require SFT, only the prompt text is needed, making the data curation process text-only and considerably simpler than collecting paired text–image data.

To validate universal applicability of our method, we experiment on three families of large language models: Llama (3.3-70B-Instruct, 3.1-8B-Instruct, 3.2-3B-Instruct)~\cite{dubey2024llama}, DeepSeek-R1-Distill 70B~\cite{guo2025deepseek}, and Qwen3 32B~\cite{yang2025qwen3}, as well as four representative text-to-image generation backbones: Flux.1-Schnell and Flux.1-Dev~\cite{flux2024}, Stable Diffusion 3.5 Medium~\cite{esser2024scaling}, and JanusPro~\cite{chen2025janus}. 

\subsection{Evaluation Metrics and Benchmarks}



We employ both standard benchmarks and multimodal LLM-based judgment. Image quality is assessed using the Fréchet Inception Distance (FID) on MS-COCO-30K~\cite{lin2014microsoft}, following the protocol of Pavlov et al.~\cite{boomb0omT2IBenchmark}. Text–image alignment is evaluated on GenEval~\cite{ghosh2023geneval}, the T2I-CompBench++~\cite{huang2025t2i} test split, and the TIFA Benchmark~\cite{hu2023tifa}. Aesthetics is measured with a LAION-based predictor~\cite{schuhmann2022laion}. To complement these automated metrics, we further employ GPT-4o as a multimodal LLM-as-a-judge on the Pick-a-Pic v2~\cite{kirstain2023pick} test split and PartiPrompts~\cite{DBLP:journals/tmlr/YuXKLBWVKYAHHPLZBW22}, where it provides pairwise ratings for image quality, text–image alignment, and aesthetics. 

\subsection{Settings and Hyperparameters}

The models were trained on 64 NVIDIA H100-SXM GPUs, with the training parameters summarized in Table~\ref{tab:training_params}. For image generation with T2I models, a resolution of 1024$\times$1024 was used for FLUX.1-schnell, FLUX.1-dev, and SD-3.5-medium, while a resolution of 384$\times$384 was used for the JanusPro experiments. The default generation parameters for T2I were applied.

\subsection{Baselines}

We compare \methodname\ with multiple baselines:
\begin{itemize}
    \item \textbf{In-context learning (ICL)}: We prompt the model with a set of curated examples and detailed prompts to rewrite the model.
    \item \textbf{Idea2Img}~\cite{yang2023idea2img} employs GPT-4V to iteratively refine prompts through feedback, memory, and draft image selection, thereby enhancing automatic image design.
    \item \textbf{PARM}~\cite{DBLP:journals/corr/abs-2501-13926} performs adaptive step-wise evaluation. It introduces three tasks: (1) clarity judgment to skip blurry early steps, (2) potential assessment to prune low-potential paths once images are sufficiently clear, and (3) best-of-N selection across high-potential candidates
    \item \textbf{PARM++}~\cite{DBLP:journals/corr/abs-2501-13926} extends PARM with a reflection mechanism. After candidate selection, it performs reflection evaluation to check for image-text misalignments. If discrepancies are detected (e.g., wrong objects, colors, or positions), PARM++ generates explicit reasons and instructs the generative model to iteratively self-correct its outputs until alignment is achieved.
    \item \textbf{Reprompt}~\cite{wu2025reprompt} trains a language model to produce structured, reasoning-driven prompts via GRPO. We discuss the detailed differences between Reprompt and our work in Sec.~\ref{appendix:diff}.
\end{itemize}

\begin{table}[]
\centering
\caption{Training parameters in our experiments.}
\label{tab:training_params}
\begin{tabular}{@{}lc@{}}
\toprule
Parameter Name & Value \\ \midrule
Prompt Rewriter temperature & 1.5 \\
DPO $\beta$& 0.1 \\
maximum iterative DPO round & 6 \\
samples-per-round $m$ & 10000 \\
candidates-per-prompt $n$ & 5 \\
epochs per DPO round & 5 \\
lora rank & 64 \\
DPO batch size & 256 \\
learning rate & 5.0e-6 \\
lr scheduler & cosine \\
warmup ratio & 0.1 \\
optimizer & AdamW \\
AdamW $\beta$ & (0.9, 0.999) \\
AdamW $\epsilon$ & 1e-8 \\
AdamW weight decay & 0.01 \\ \bottomrule
\end{tabular}
\end{table}

\section{Differences with RePrompt}
\label{appendix:diff}

In this section, we provide a detailed comparison between \methodname\ and the concurrent work Reprompt~\cite{wu2025reprompt}.

From the algorithmic perspective, our approach does not include an SFT stage. Our preliminary experiments revealed that SFT constrains the exploration ability of LLMs, leading to overfitting on the SFT training data. Moreover, collecting diverse, high-quality, and unbiased training data tailored to each T2I model is both challenging and costly. By directly applying RL for training, our method only requires raw user prompts as inputs, without the need for model-specific datasets.

Another key distinction lies in the optimization paradigm. Reprompt adopts GRPO, whereas we employ DPO. We argue that reward estimation for generated images is inherently difficult under GRPO, while the construction of chosen and rejected prompt pairs is relatively easy. This enables more reliable reward guidance in DPO. Consistent with our findings, Tong et al.~\cite{tong2025delving} also report that DPO achieves superior in-domain performance compared to GRPO.

In terms of reward design, Reprompt primarily focuses on text–image alignment and only reports results on GenEval and T2I-Compbench, both of which emphasize alignment evaluation. However, no results are provided for image quality or aesthetics. In contrast, our framework evaluates these additional dimensions by reporting image quality and aesthetics metrics on Pick-a-Pic v2 and PartiPrompts, as well as FID scores on MS COCO 30k. Our experiments further reveal a clear trade-off among alignment, quality, and aesthetics, highlighting the necessity of improving alignment without compromising other aspects.

From the experimental perspective, we comprehensively evaluated the scalability and transferability. Whereas Reprompt is limited to Qwen2.5-3B, we incorporate a broader range of LLMs and T2I model backbones, and also conduct large-scale experiments with 70B-parameter LLMs. 

Finally, \methodname\ consistently outperforms Reprompt on GenEval and T2I Compbench, while simultaneously advancing performance on image quality and aesthetics benchmarks.

\section{Additional Experiments}




\subsection{Transferability}

We further evaluate the transferability on the GenEval datasets in Table~\ref{tab:geneval_transfer}. The results show that, even when the training and testing T2I backbones differ, \methodname\ still achieves good performance, demonstrating robust transferability.
\begin{table*}[]
\centering
\caption{Transferability results evaluated on GenEval. We train rewriters on different T2I backbones and evaluate on fixed target backbones (FLUX.1-schnell and SD-3.5-medium). The results demonstrate that, even when the training and testing backbones differ, our method still achieves significant performance improvements. ``()'' shows what T2I model is used during training.}
\label{tab:geneval_transfer}
\setlength{\tabcolsep}{2pt}
\resizebox{\textwidth}{!}{%
\begin{tabular}{@{}lccccccc@{}}
\toprule
\multicolumn{1}{l|}{Inference T2I model} & \begin{tabular}[c]{@{}c@{}}Single\\ object\end{tabular} & \begin{tabular}[c]{@{}c@{}}Two\\ object\end{tabular} & Counting & Colors & Position & \multicolumn{1}{c|}{\begin{tabular}[c]{@{}c@{}}Color\\ attribution\end{tabular}} & Overall \\ \midrule
\rowcolor{blue!20}\multicolumn{8}{c}{Training T2I Backbone for \methodname: FLUX.1-schnell} \\ \midrule
\multicolumn{1}{l|}{FLUX.1-schnell (orig. prompt)} & 0.99 & 0.87 & 0.61 & 0.79 & 0.35 & \multicolumn{1}{c|}{0.43} & 0.67 \\ \midrule
\multicolumn{1}{l|}{FLUX.1-schnell (+\methodname)} & 0.99 & 0.91 & 0.65 & 0.77 & \textbf{0.64} & \multicolumn{1}{c|}{\textbf{0.57}} & 0.75 \\
\multicolumn{1}{l|}{FLUX.1-dev (+\methodname)} & 0.99 & 0.93 & 0.66 & \textbf{0.86} & 0.55 & \multicolumn{1}{c|}{\textbf{0.57}} & 0.76 \\
\multicolumn{1}{l|}{SD-3.5-medium (+\methodname)} & \textbf{1.00} & \textbf{0.95} & \textbf{0.72} & 0.81 & 0.61 & \multicolumn{1}{c|}{0.66} & \textbf{0.79} \\
\multicolumn{1}{l|}{JanusPro(+\methodname)} & 0.99 & 0.87 & 0.63 & 0.81 & 0.56 & \multicolumn{1}{c|}{\textbf{0.57}} & 0.74 \\ \midrule
\rowcolor{blue!20}\multicolumn{8}{c}{Training T2I Backbone for \methodname: SD-3.5-medium} \\ \midrule
\multicolumn{1}{l|}{SD-3.5-medium (orig. prompt)} & \textbf{1.00} & 0.84 & 0.68 & 0.85 & 0.24 & \multicolumn{1}{c|}{0.61} & 0.70 \\ \midrule
\multicolumn{1}{l|}{FLUX.1-schnell (+\methodname)} & \textbf{1.00} & \textbf{0.92} & \textbf{0.69} & 0.83 & \textbf{0.58} & \multicolumn{1}{c|}{\textbf{0.69}} & \textbf{0.78} \\
\multicolumn{1}{l|}{FLUX.1-dev (+\methodname)} & \textbf{1.00} & \textbf{0.92} & 0.68 & \textbf{0.88} & 0.51 & \multicolumn{1}{c|}{0.60} & 0.76 \\
\multicolumn{1}{l|}{SD-3.5-medium (+\methodname)} & \textbf{1.00} & \textbf{0.92} & 0.68 & 0.85 & 0.57 & \multicolumn{1}{c|}{0.56} & 0.76 \\
\multicolumn{1}{l|}{JanusPro(+\methodname)} & \textbf{1.00} & 0.88 & 0.70 & 0.86 & 0.54 & \multicolumn{1}{c|}{0.54} & 0.75 \\ \bottomrule
\end{tabular}
}
\end{table*}


\subsection{Performance of JanusPro on GenEval}

We report counterintuitive results for JanusPro on GenEval in Table~\ref{tab:januspro_geneval}, where both ICL and our method fail to yield improvements. In contrast, our approach achieves performance gains for JanusPro on T2I-CompBench++ and TIFA-Benchmark. Interestingly, JanusPro outperforms the other three diffusion-based models on GenEval, yet performs consistently worse on T2I-CompBench++ and TIFA-Benchmark. We hypothesize that this discrepancy arises from JanusPro potentially overfitting to the GenEval benchmark.

\begin{table*}[]
\centering
\caption{GenEval results for JanusPro, where both ICL and our method do not yield performance improvements.}
\label{tab:januspro_geneval}
\begin{tabular}{@{}l|cccccc|c@{}}
\toprule
Model & \begin{tabular}[c]{@{}c@{}}Single\\ object\end{tabular} & \begin{tabular}[c]{@{}c@{}}Two\\ object\end{tabular} & Counting & Colors & Position & \begin{tabular}[c]{@{}c@{}}Color\\ attribution\end{tabular} & Overall \\ \midrule
JanusPro & \textbf{1.00} & 0.86 & \textbf{0.59} & \textbf{0.89} & \textbf{0.73} & \textbf{0.65} & \textbf{0.79} \\
+ ICL & \textbf{1.00} & \textbf{0.93} & 0.39 & 0.86 & 0.43 & 0.48 & 0.68 \\
+ \methodname & \textbf{1.00} & 0.81 & 0.38 & 0.86 & 0.60 & 0.43 & 0.68 \\ \bottomrule
\end{tabular}
\end{table*}

\section{Ablation Study}

\subsection{Reward Type}

Table~\ref{tab:ablation} presents an ablation study of the reward terms. Removing the quality reward substantially decreases the image–quality win rate, from 0.494 to 0.364. Excluding either the general or physical alignment rewards reduces alignment, from 0.561 to 0.521 and to 0.504 respectively, confirming that each reward effectively drives its intended objective. Incorporating the aesthetics reward yields a pronounced improvement in aesthetics, increasing from 0.476 to 0.818, but simultaneously reduces alignment from 0.561 to 0.424, highlighting a trade-off. Qualitative analysis in Table~\ref{tab:aes_case_study} suggests that the aesthetics reward often enriches scenes by introducing additional objects or ornamentation, which can dilute the main subject and thereby weaken alignment. To accommodate different objectives, we report two variants: a general rewriter optimized for faithfulness and overall quality, and an aesthetics rewriter tailored to visual appeal.

\begin{table*}[]
\centering
\caption{Reward ablation results for \methodname. Removing any individual reward leads to a decline in its associated metric. Incorporating the aesthetics reward improves aesthetic preference but decreases alignment, thereby revealing an inherent trade-off.}
\label{tab:ablation}
\resizebox{\textwidth}{!}{%
\begin{tabular}{@{}l|c|cccc@{}}
\toprule
 & Rewriter Type & \begin{tabular}[c]{@{}c@{}}Image\\ Quality\end{tabular} & \begin{tabular}[c]{@{}c@{}}Image\\ Aesthetics\end{tabular} & \begin{tabular}[c]{@{}c@{}}Image-Text\\ Alignment\end{tabular} & Average \\ \midrule
FLUX.1-schnell & - & 0.469 & 0.314 & 0.419 & 0.401 \\
+ ICL & - & 0.475 & 0.307 & 0.422 & 0.401 \\
+ Zero Shot & - & 0.431 & 0.263 & 0.417 & 0.370 \\
+ \methodname & General Rewriter, w/o quality reward & 0.364 & 0.461 & \textbf{0.563} & 0.463 \\
+ \methodname & General Rewriter, w/o general alignment reward & \underline{0.523} & 0.470 & 0.521 & 0.505 \\
+ \methodname & General Rewriter, w/o physical alignment reward & \textbf{0.533} & 0.448 & 0.504 & 0.495 \\
+ \methodname & General Rewriter & 0.494 & 
\underline{0.476} & \underline{0.561} & \underline{0.510} \\
+ \methodname & Aesthetics Rewriter & 0.495 & \textbf{0.818} & 0.424 & \textbf{0.579}\\ \bottomrule
\end{tabular}%
}
\end{table*}

\subsection{Performance for each round}

\begin{figure*}[htbp]
    \centering

    \begin{subfigure}[tb]{0.32\textwidth}
        \centering
        \includegraphics[width=\linewidth]{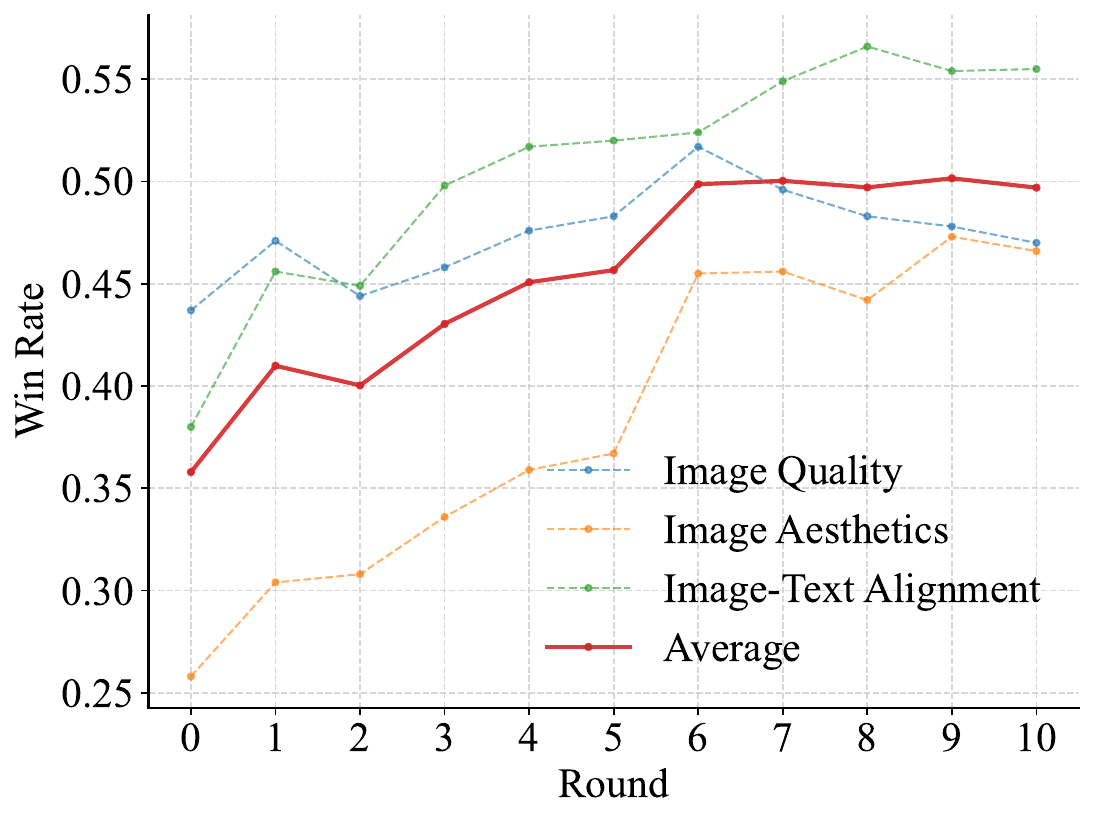}
        \caption{Llama3.2-3B-Instruct}
    \end{subfigure}
    \hfill
    \begin{subfigure}[tb]{0.32\textwidth}
        \centering
        \includegraphics[width=\linewidth]{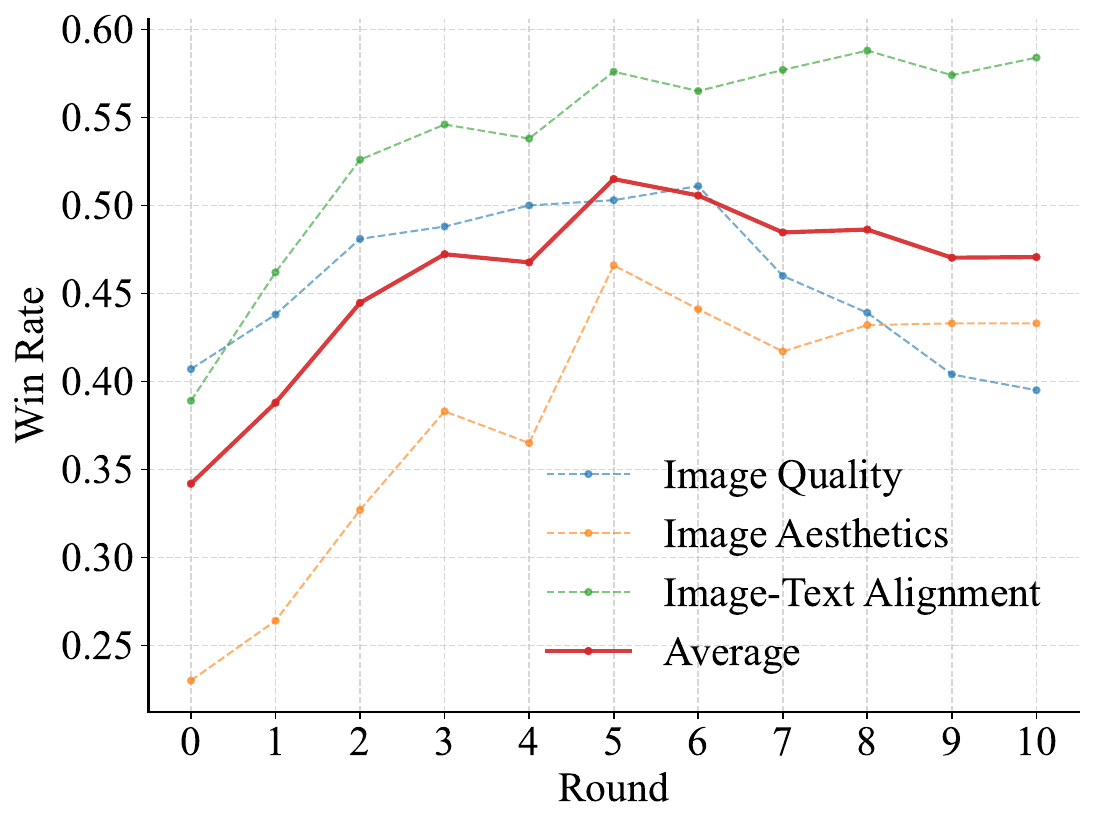}
        \caption{Llama3.1-8B-Instruct}
    \end{subfigure}
    \hfill
    \begin{subfigure}[tb]{0.32\textwidth}
        \centering
        \includegraphics[width=\linewidth]{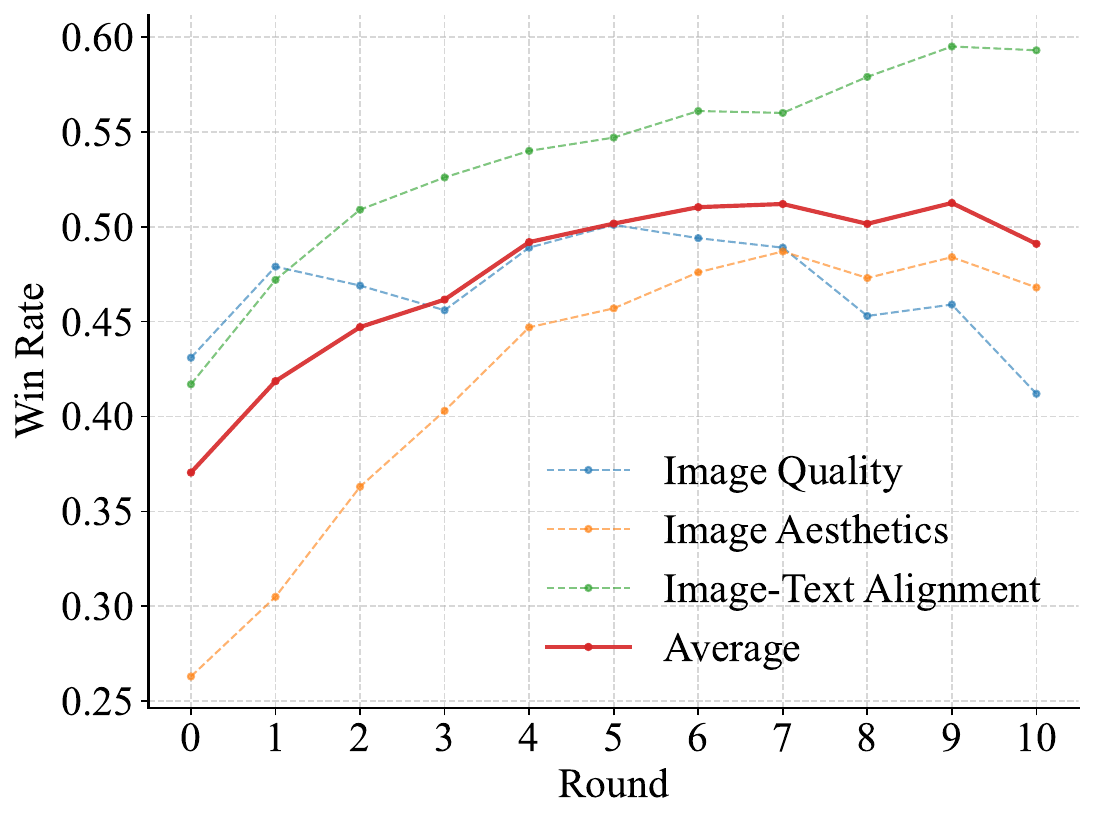}
        \caption{Llama3.3-70B-Instruct}
    \end{subfigure}
    
    \caption{Win rate for different base models in different rounds. The prompt rewriters are evaluated on Pick-a-pic v2 with GPT-4o.}
    \label{fig:exp_round}
\end{figure*}

We report the win rates of different base models across DPO training rounds in Table~\ref{fig:exp_round}. During the first six rounds, all metrics exhibit steady improvement. Beyond round six, however, image quality and aesthetics begin to decline, while text–image alignment continues to improve, reflecting an inherent trade-off. Consequently, the average win rate decreases, indicating that this trade-off becomes more pronounced after six rounds. It should also be noted that our training dataset contains approximately 60k samples, with 10k used per round; thus, the first six rounds effectively cover the entire dataset once. This observation suggests that expanding the dataset could yield further gains, and that a single iteration over the training data is sufficient, whereas additional iterations may lead to overfitting and degraded performance.

\subsection{Rewritten Prompt Length}
\label{Appendix:rewritten_length}

\begin{figure*}
    \centering
    \includegraphics[width=0.4\linewidth]{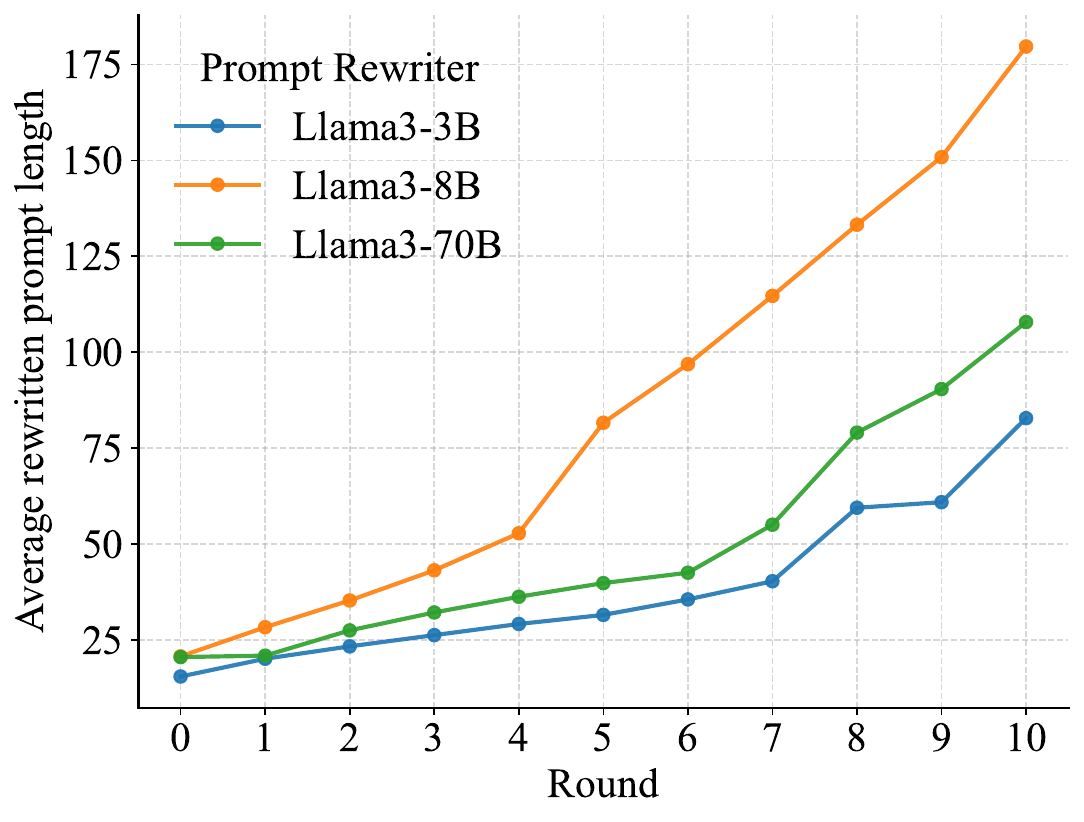}
    \caption{Average length of rewritten prompts across different DPO training rounds, computed on the Pick-a-Pic v2 dataset.}
    \label{fig:rewritten_length}
\end{figure*}

Figure~\ref{fig:rewritten_length} illustrates a steady increase in the average token length of rewritten prompts over training. The rewriter progressively incorporates additional specificity, such as attributes, relations, and constraints, and this increase in length correlates with higher win rates.

\subsection{Full Finetune vs. LoRA}



We conduct a controlled comparison between LoRA and full-model finetuning under identical experimental conditions, as shown in Figure~\ref{fig:fft_vs_lora}. Full-model finetuning does not yield consistent improvements, whereas LoRA attains comparable performance while requiring substantially less memory and computation. These results suggest that LoRA constitutes a strong and practical choice for prompt-rewriter training.

\section{Case Study}
\label{appendix:case_study}

\subsection{General Rewriter}

We provide five illustrative examples for the general rewriter in Tables~\ref{tab:case1},~\ref{tab:case2},~\ref{tab:case3},~\ref{tab:case4}, and~\ref{tab:case5}. In these examples, we employ Llama3-70B-Instruct as the prompt rewriter and FLUX.1-schnell as the backbone T2I model. To highlight the training dynamics, we present the rewritten prompts together with their corresponding generated images across successive DPO training rounds, where round 0 corresponds to the original user-provided prompt. As training progresses, the rewritten prompts become increasingly detailed and precise, resulting in improvements in image quality, text–image alignment, and aesthetics.

\subsection{Aesthetics Rewriter}

We also observe that the aesthetics rewriter can, in some cases, compromise image–text alignment relative to the general rewriter, as illustrated by the examples in Table~\ref{tab:aes_case_study}.

\begin{table*}[]
\centering
\caption{Case study example 1.}
\label{tab:case1}
\begin{tabular}{ >{\centering\arraybackslash}m{1cm} | >{\centering\arraybackslash}m{8cm} | >{\centering\arraybackslash}m{2cm} }
\toprule
Round & \multicolumn{1}{c|}{Prompt} & Image \\ \midrule
0 & city skyline in visual novel & \includegraphics[width=0.15\textwidth]{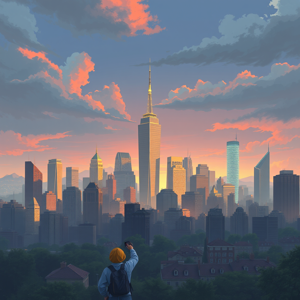} \\ \midrule
1 & \begin{tabular}[c]{@{}l@{}}A vibrant cityscape serves as the backdrop in this \\ visual novel, with a sprawling metropolis skyline\\  featuring towering skyscrapers and bustling streets.\end{tabular} & \includegraphics[width=0.15\textwidth]{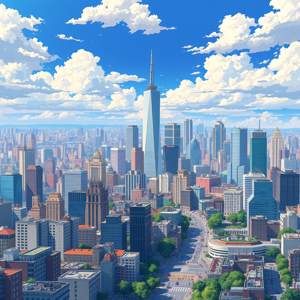} \\ \midrule
2 & \begin{tabular}[c]{@{}l@{}}A vibrant cityscape serves as the backdrop for a\\  dramatic scene in a visual novel, with towering\\  skyscrapers and neon lights illuminating the \\ night sky.\end{tabular} & \includegraphics[width=0.15\textwidth]{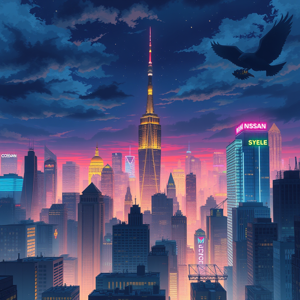} \\ \midrule
3 & \begin{tabular}[c]{@{}l@{}}A futuristic cityscape serves as the backdrop for a\\  dramatic scene in a visual novel, with sleek \\ skyscrapers and neon lights illuminating the night\\  sky as characters' dialogue and choices hang in the\\  balance.\end{tabular} & \includegraphics[width=0.15\textwidth]{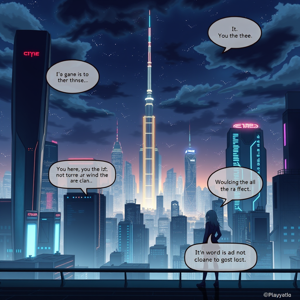} \\ \midrule
4 & \begin{tabular}[c]{@{}l@{}}A futuristic cityscape at dusk, with sleek skyscrapers \\ and neon lights illuminating the horizon, serves as \\ the backdrop for a dramatic scene in a Japanese-style \\ visual novel, complete with dialogue bubbles and \\ character sprites.\end{tabular} & \includegraphics[width=0.15\textwidth]{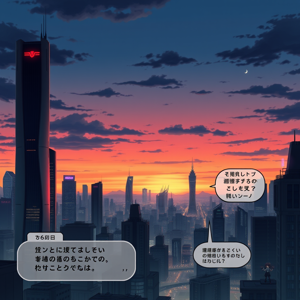} \\ \midrule
5 & \begin{tabular}[c]{@{}l@{}}A futuristic cityscape at dusk serves as the backdrop \\ for a visual novel scene, with sleek skyscrapers \\ and neon-lit billboards towering above, as characters \\ in the foreground engage in a dramatic \\ conversation amidst the bustling metropolis.\end{tabular} & \includegraphics[width=0.15\textwidth]{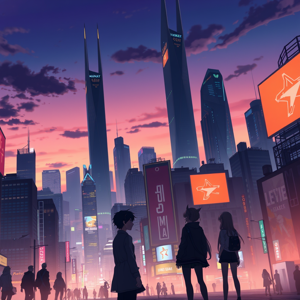} \\ \midrule
6 & \begin{tabular}[c]{@{}l@{}}A futuristic cityscape at dusk serves as the backdrop \\ for a visual novel interface, featuring a ornate user\\  interface with anime-style characters and dialogue \\ boxes amidst a sprawling metropolis with neon-lit \\ skyscrapers, bustling streets, and a vibrant night sky.\end{tabular} & \includegraphics[width=0.15\textwidth]{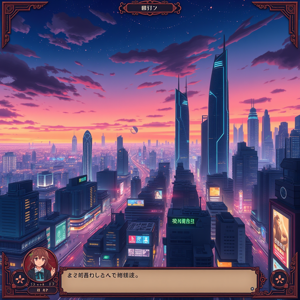} \\ \bottomrule
\end{tabular}
\end{table*}

\begin{table*}[]
\centering
\caption{Case study example 2.}

\label{tab:case2}
\begin{tabular}{ >{\centering\arraybackslash}m{1cm} | >{\centering\arraybackslash}m{8cm} | >{\centering\arraybackslash}m{2cm} }
\toprule
Round & \multicolumn{1}{c|}{Prompt} & Image \\ \midrule
0 & Movie poster artwork of a four-armed cyborg alien. & \includegraphics[width=0.15\textwidth]{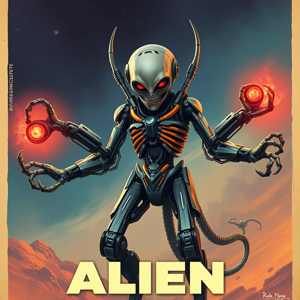} \\ \midrule
1 & \begin{tabular}[c]{@{}l@{}}Vibrant movie poster featuring a futuristic, \\ four-armed cyborg with an otherworldly alien physique.\end{tabular} & \includegraphics[width=0.15\textwidth]{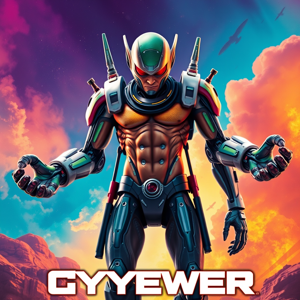} \\ \midrule
2 & \begin{tabular}[c]{@{}l@{}}Vibrant movie poster featuring a futuristic cyborg \\ alien with four powerful arms, blending human and \\ extraterrestrial elements in a stunning sci-fi design.\end{tabular} & \includegraphics[width=0.15\textwidth]{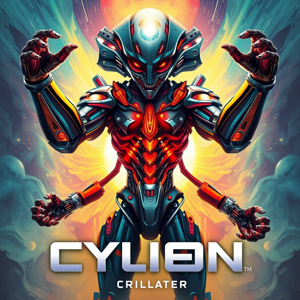} \\ \midrule
3 & \begin{tabular}[c]{@{}l@{}}A futuristic movie poster featuring a powerful cyborg \\ alien with four arms, surrounded by a vibrant, \\ otherworldly landscape and explosive action elements.\end{tabular} & \includegraphics[width=0.15\textwidth]{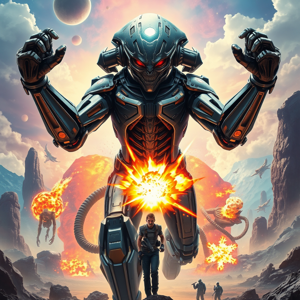} \\ \midrule
4 & \begin{tabular}[c]{@{}l@{}}A futuristic movie poster featuring a menacing \\ four-armed cyborg alien with a metallic exoskeleton\\  and glowing blue circuits, standing amidst a backdrop\\  of exploding spaceships and a war-torn planetary\\  landscape.\end{tabular} & \includegraphics[width=0.15\textwidth]{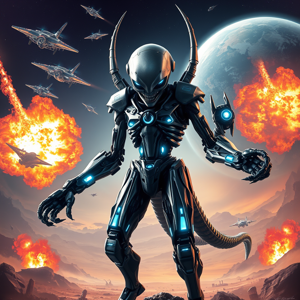} \\ \midrule
5 & \begin{tabular}[c]{@{}l@{}}A sci-fi movie poster featuring a futuristic cyborg\\  alien with four arms, surrounded by a cosmic \\ landscape and advanced technology, with the\\  title and credits of the film emblazoned in\\  bold, neon-lit letters.\end{tabular} & \includegraphics[width=0.15\textwidth]{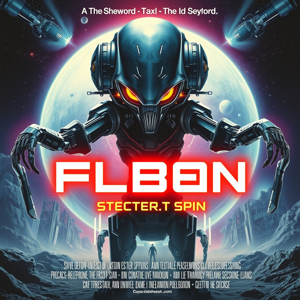} \\ \midrule
6 & \begin{tabular}[c]{@{}l@{}}A sci-fi movie poster featuring a futuristic cyborg\\  alien with four arms, adorned in metallic armor\\  and surrounded by a cosmic landscape, with\\  the title and credits of the film emblazoned \\ in bold, neon-lit letters.\end{tabular} & \includegraphics[width=0.15\textwidth]{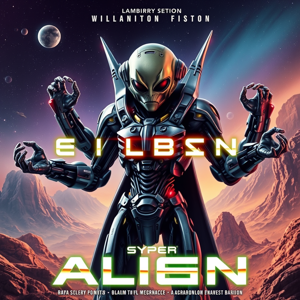} \\ \bottomrule
\end{tabular}
\end{table*}

\begin{table}[]
\centering
\caption{Case study example 3.}
\label{tab:case3}
\renewcommand{\arraystretch}{1.5} 
\tiny{
\begin{tabular}{|p{1cm}|p{8cm}|p{2cm}}
\toprule
Round & \multicolumn{1}{c|}{Prompt} & Image \\ \midrule
0 & full body, a slender anime girl with long cyan hair, powerful arcane wizard, beautiful white outfit, extremely detailed, realistic shading & \vspace{0pt} \includegraphics[width=0.15\textwidth]{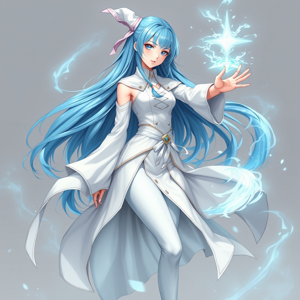} \\ \midrule
1 & A stunning, slender anime girl with flowing cyan locks stands tall, radiating magical prowess as a powerful arcane wizard. She wears a breathtakingly beautiful white outfit, showcased in exquisite detail with remarkably realistic shading. & \vspace{0pt}\includegraphics[width=0.15\textwidth]{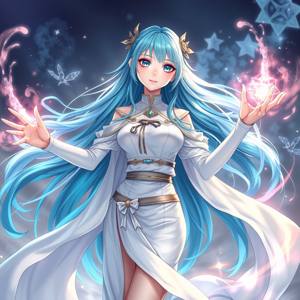} \\ \midrule
2 & A stunning, slender anime girl stands tall, her lengthy cyan locks cascading down her back as she embodies the essence of a powerful arcane wizard. Adorned in a breathtakingly beautiful, intricately designed white outfit, she exudes magic and sophistication. The artwork boasts extremely detailed textures and realistic shading, bringing this enchanting scene to life. & \vspace{0pt}\includegraphics[width=0.15\textwidth]{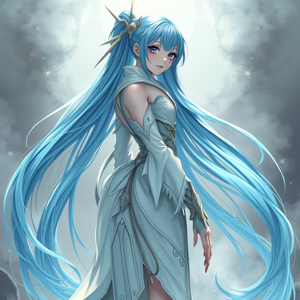} \\ \midrule
3 & A stunning anime sorceress stands tall, clad in an exquisite white ensemble, as her slender figure is surrounded by an aura of mystical energy. Her lengthy cyan locks cascade down her back like a river of ice, contrasting beautifully with the intricate, ornate details of her attire. With a commanding presence, she wields her magical prowess, as the atmosphere is electrified by her arcane abilities. The artwork boasts extremely realistic shading, bringing every facet of this enchanting scene to life with breathtaking precision. & \vspace{0pt}\includegraphics[width=0.15\textwidth]{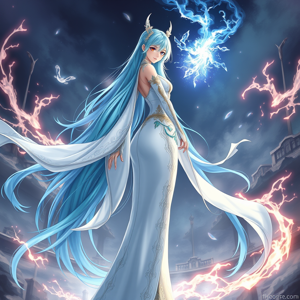} \\ \midrule
4 & A majestic, slender anime sorceress stands tall, adorned in a exquisite, intricately designed white attire, surrounded by a halo of mystical energy. Her long, flowing cyan hair cascades down her back like a river of celestial magic. With an ornate staff in hand, she channels  powerful arcane forces, as evidenced by the swirling, ethereal auras and glowing runes that dance around her. The entire scene is rendered in breathtakingly detailed,  hyper-realistic art, with masterful shading that imbues the image with depth, dimension,  and an aura of wonder. & \vspace{0pt}\includegraphics[width=0.15\textwidth]{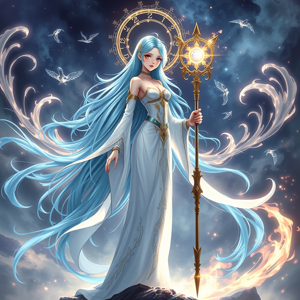} \\ \midrule
5 & A majestic, slender anime sorceress stands tall, adorned in an exquisite, intricately designed white attire, surrounded by a halo of mystical energy. Her long, flowing cyan hair cascades down her back like a river of night sky, as she wields a ornate staff crackling with arcane power. Every aspect of her regal presence is rendered in breathtaking,  hyper-realistic detail, from the delicate folds of her garments to the subtle, luminous glow of her magical aura, all set against a backdrop of subtle, gradient shading that imbues the entire scene with an aura of wonder and enchantment. & \vspace{0pt}\includegraphics[width=0.15\textwidth]{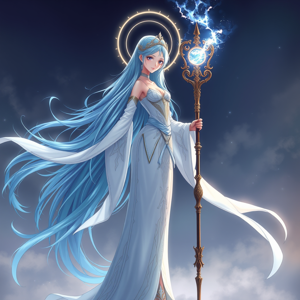} \\ \midrule
6 & A majestic, slender anime sorceress stands amidst a mystical aura, adorned in an exquisite,  intricately designed white attire adorned with golden accents and ornate gemstones, as she summons a vortex of magical energy with her staff. Her lengthy, vibrant cyan hair  cascades down her back like a waterfall, with strands flowing upwards, surrounded by a halo of radiant, swirling runes. She poses amidst a lavish, ornate backdrop of ancient tomes, mystical artifacts, and celestial bodies, surrounded by a kaleidoscope of magical orbs, stars, and glittering, ethereal particles. The atmosphere is alive with dynamic,  shimmering lights and intense, realistic shading, accentuating the intricate details of her elaborate costume, the ornate environment, and the fantastical, dreamlike scenery. & \vspace{0pt}\includegraphics[width=0.15\textwidth]{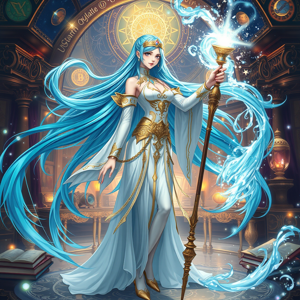} \\ \bottomrule
\end{tabular}
}
\end{table}

\begin{table*}[]
\centering
\caption{Case study example 4.}
\label{tab:case4}
\tiny{
\begin{tabular}{ >{\centering\arraybackslash}m{1cm} | >{\centering\arraybackslash}m{7cm} | >{\centering\arraybackslash}m{2cm} }
\toprule
Round & \multicolumn{1}{c|}{Prompt} & Image \\ \midrule
0 & An stylized entrance to a rocky cave & \includegraphics[width=0.15\textwidth]{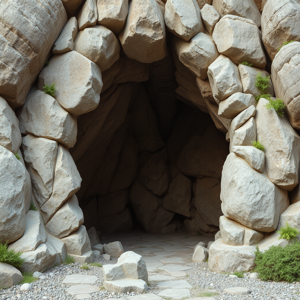} \\ \midrule
1 & A whimsical, artistic gateway to a rugged, rocky cavern. & \includegraphics[width=0.15\textwidth]{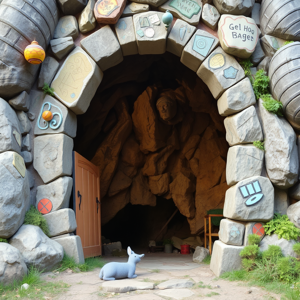} \\ \midrule
2 & \begin{tabular}[c]{@{}l@{}}A fantastical, ornate gateway leads to a mystical rocky cave, \\ surrounded by ancient stones and lush greenery.\end{tabular} & \includegraphics[width=0.15\textwidth]{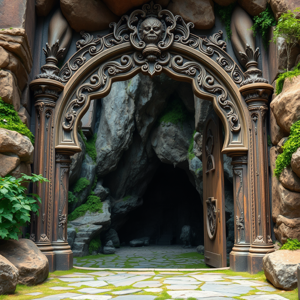} \\ \midrule
3 & \begin{tabular}[c]{@{}l@{}}A fantastical, ornate gateway leading to a mystical rocky cave,\\  surrounded by lush greenery and vines, with ancient ruins \\ and mysterious artifacts scattered throughout the entrance.\end{tabular} & \includegraphics[width=0.15\textwidth]{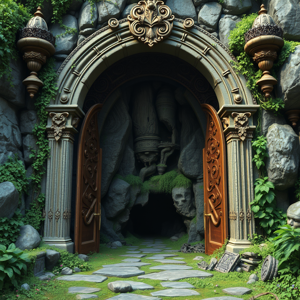} \\ \midrule
4 & \begin{tabular}[c]{@{}l@{}}A fantastical, ornate gateway adorned with mystical symbols\\  and vines, leading to a majestic, ancient rocky cave surrounded\\  by lush greenery and towering trees, with a warm, golden\\  light emanating from within.\end{tabular} & \includegraphics[width=0.15\textwidth]{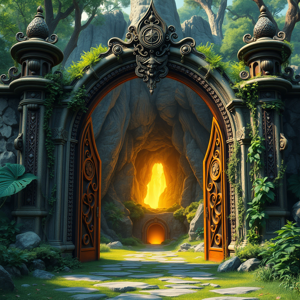} \\ \midrule
5 & \begin{tabular}[c]{@{}l@{}}A fantastical, ornate gateway adorned with ancient carvings \\ and mystical symbols, leading to a majestic, torch-lit rocky\\  cave surrounded by towering stalactites and stalagmites,\\  with a hint of misty aura and a starry night sky visible\\  in the background.\end{tabular} & \includegraphics[width=0.15\textwidth]{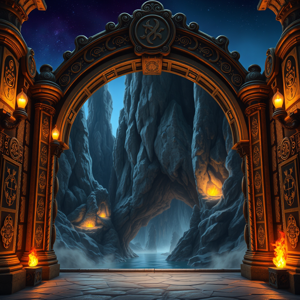} \\ \midrule
6 & \begin{tabular}[c]{@{}l@{}}A ornate, fantastical entrance adorned with ancient carvings\\  and mystical symbols, leading to a majestic rocky cave\\  surrounded by lush greenery and towering mountains in\\  the background, with a warm glow emanating from within.\end{tabular} & \includegraphics[width=0.15\textwidth]{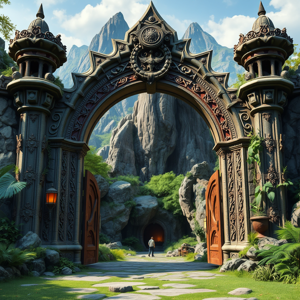} \\ \bottomrule
\end{tabular}
}
\end{table*}

\begin{table*}[]
\centering
\caption{Case study example 5.}
\label{tab:case5}
\begin{tabular}{ >{\centering\arraybackslash}m{1cm} | >{\centering\arraybackslash}m{7cm} | >{\centering\arraybackslash}m{2cm} }
\toprule
Round & \multicolumn{1}{c|}{Prompt} & Image \\ \midrule
0 & an anime girl in a kimono & \includegraphics[width=0.15\textwidth]{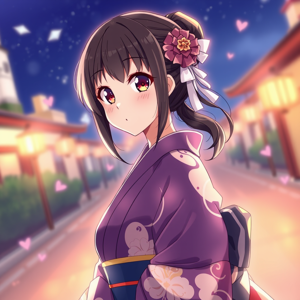} \\ \midrule
1 & \begin{tabular}[c]{@{}l@{}}A young woman with Japanese-inspired features\\  and styling wears a traditional kimono, adorned\\  with intricate designs and patterns, in a vibrant\\  anime art style.\end{tabular} & \includegraphics[width=0.15\textwidth]{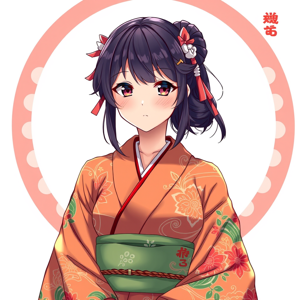} \\ \midrule
2 & \begin{tabular}[c]{@{}l@{}}A beautiful anime girl adorned in a vibrant \\ traditional Japanese kimono, surrounded by\\  a serene atmosphere.\end{tabular} & \includegraphics[width=0.15\textwidth]{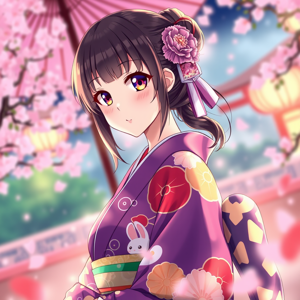} \\ \midrule
3 & \begin{tabular}[c]{@{}l@{}}A beautiful anime girl adorned in a traditional\\  Japanese kimono, surrounded by a serene\\  atmosphere, with intricate designs and patterns\\  on her garment, and a gentle smile on her face.\end{tabular} & \includegraphics[width=0.15\textwidth]{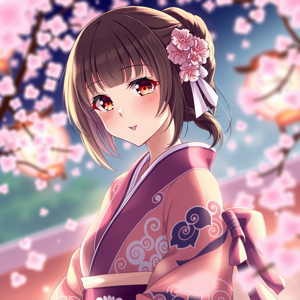} \\ \midrule
4 & \begin{tabular}[c]{@{}l@{}}A beautiful anime girl adorned in a traditional\\  Japanese kimono, surrounded by cherry\\  blossoms, standing in a serene garden setting.\end{tabular} & \includegraphics[width=0.15\textwidth]{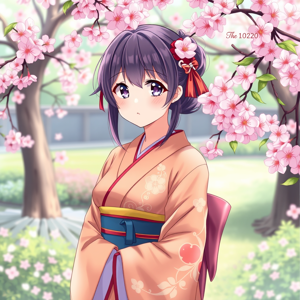} \\ \midrule
5 & \begin{tabular}[c]{@{}l@{}}A beautiful anime girl adorned in a traditional\\  Japanese kimono, surrounded by cherry\\  blossoms, standing in a serene garden with \\ a tranquil waterfall in the background.\end{tabular} & \includegraphics[width=0.15\textwidth]{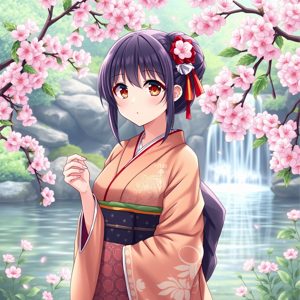} \\ \midrule
6 & \begin{tabular}[c]{@{}l@{}}A beautiful anime girl adorned in a traditional\\  Japanese kimono, surrounded by cherry\\  blossoms, posing elegantly in a serene\\  Japanese garden.\end{tabular} & \includegraphics[width=0.15\textwidth]{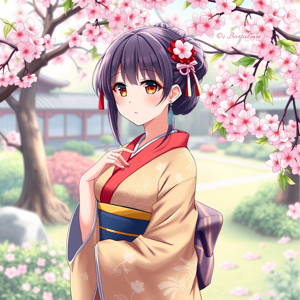} \\ \bottomrule
\end{tabular}
\end{table*}

\begin{table}[]
\centering
\caption{Case study for the aesthetic rewriter. While it often enhances visual appeal, it can sometimes lead to prompt misalignment. For example, in Image 1, the result is more visually pleasing but rendered brighter, contradicting the user’s low-light requirement. In Image 2, the rewriter overlooks the instruction to depict a fat man, producing a misaligned outcome.}
\label{tab:aes_case_study}
\tiny{
\begin{tabular}{|l|p{3.5cm}|p{3.5cm}|p{3.5cm}|}
\toprule
 & No Rewriter & General Rewriter & Aesthetics Rewriter \\ \midrule
Prompt1 & a low light photo of a city at night & A moody nighttime cityscape photograph captured in low light, depicting a bustling metropolis adorned with twinkling skyscrapers, neon signs, and streets aglow with vehicle headlights, set against a dark starry sky. & A breathtaking cityscape at dusk surrounded by a kaleidoscope of twinkling lights,  majestic skyscrapers, and bustling streets, set against a backdrop of a vibrant purple and blue night sky filled with stars, captured in a stunning low-light photograph. \\
\vspace{0pt}Image1 & \includegraphics[width=0.28\textwidth]{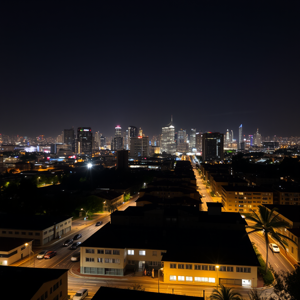} & \includegraphics[width=0.28\textwidth]{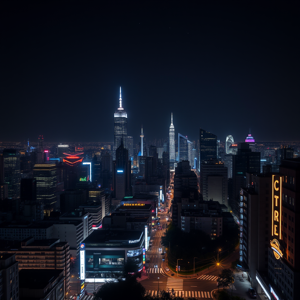} & \includegraphics[width=0.28\textwidth]{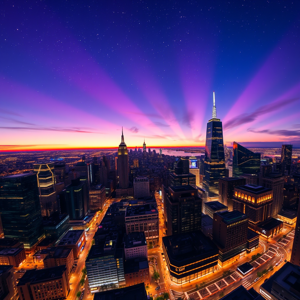} \\ \midrule
Prompt2 & Fat black man in suit watching iguanas in a petting zoo realistic hyper realistic photorealistic & A photorealistic image depicting a obese African American man dressed in a formal business suit, standing in a lushpetting zoo surrounded by various tropicalplants and trees, intently observing agroup of iguanas roaming freely aroundhim, with some of the reptiles climbingon his legs and arms, as he gently offersthem leaves to eat, amidst a backdrop of a sunny day with a clear blue sky. & A lavishly dressed black man surrounded by opulent decorations and lush greenery, sits in a luxurious golden chair, observing in wonder as a group of vibrant iguanas roam and play in a majestic petting zoo filled with exotic flowers, sparkling fountains, and intricate stone statues, surrounded by a backdrop of majestic waterfalls and a breathtaking sunset, amidst a scene of unparalleled realism, surrounded by intricate details and majestic architecture, adorned with gold ornaments and precious gems,  surrounded by a warm and vibrant ambiance,  illuminated by the golden light of the setting sun, in an incredibly detailed and intricate hyper-realistic artwork. \\
Image2 & \includegraphics[width=0.28\textwidth]{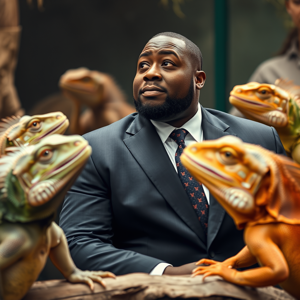} & \includegraphics[width=0.28\textwidth]{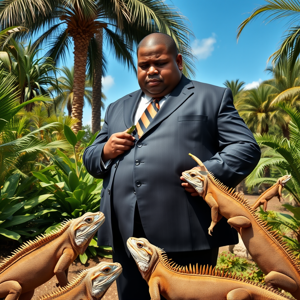} & \includegraphics[width=0.28\textwidth]{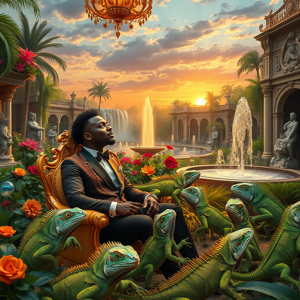} \\ \bottomrule
\end{tabular}
}
\end{table}

\begin{figure*}[h]
    \centering
    \includegraphics[width=0.8\linewidth]{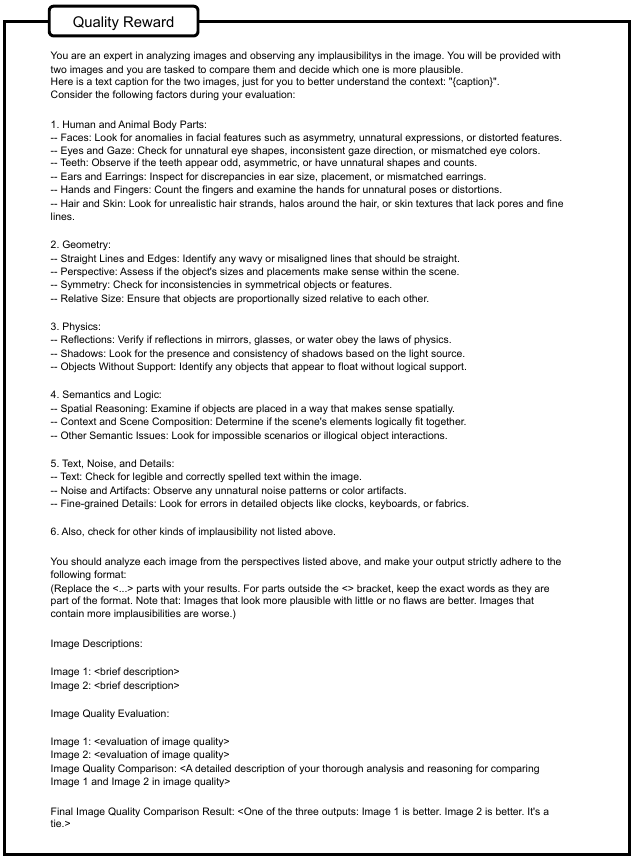}
    \caption{Prompt for quality reward $r_\text{Quality}$.}
    \label{fig:quality_reward}
\end{figure*}

\begin{figure*}[h]
    \centering
    \includegraphics[width=0.8\linewidth]{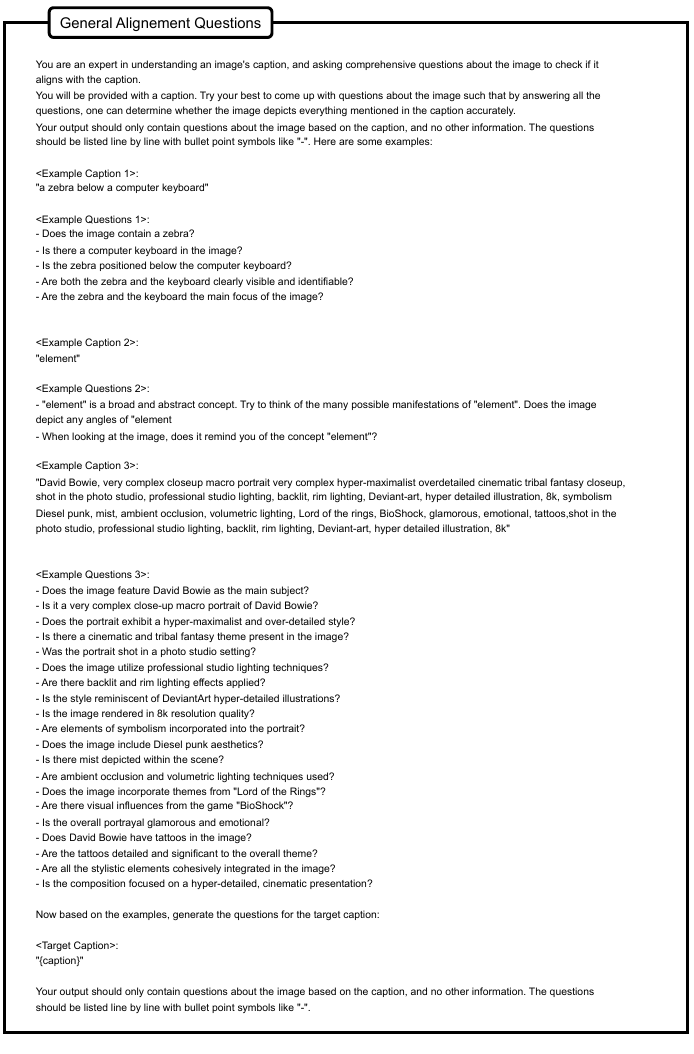}
    \caption{Prompt for general alignment questions.}
    \label{fig:general_alignment_questions}
\end{figure*}

\begin{figure*}[h]
    \centering
    \includegraphics[width=0.8\linewidth]{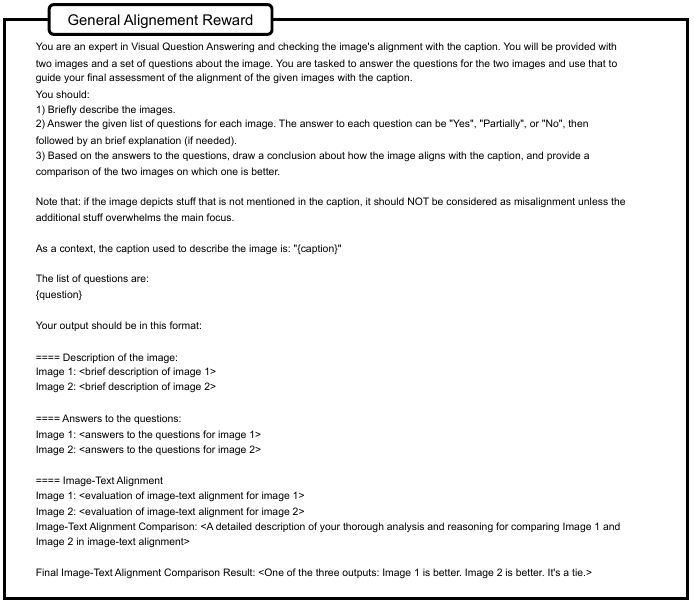}
    \caption{Prompt for general alignment reward $r_\text{General-Alignment}$.}
    \label{fig:general_alignment_reward}
\end{figure*}

\begin{figure*}[h]
    \centering
    \includegraphics[width=0.8\linewidth]{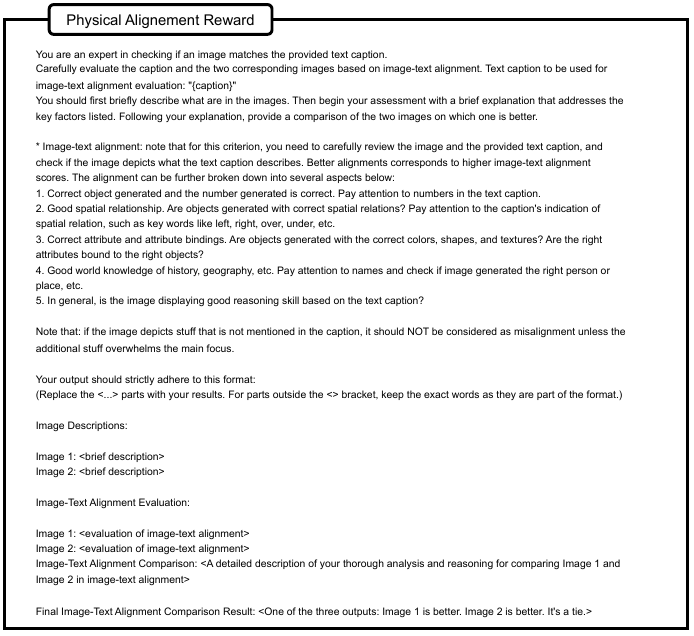}
    \caption{Prompt for physical alignment reward $r_\text{Physical-Alignment}$}
    \label{fig:physical_alignment_reward}
\end{figure*}

\begin{figure*}[h]
    \centering
    \includegraphics[width=0.8\linewidth]{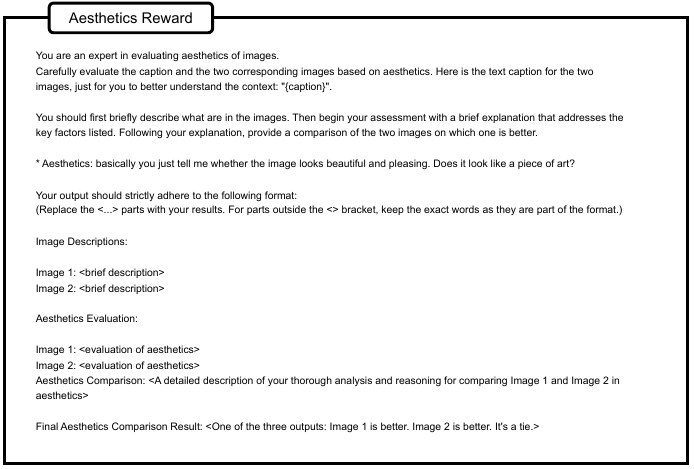}
    \caption{Prompt for aesthetics reward $r_\text{Aesthetics}$}
    \label{fig:aesthetics_reward}
\end{figure*}

\section{Detailed Reward Design}
\label{appendix:reward}

In this section, we present the prompts employed for reward evaluation, corresponding to Figures~\ref{fig:quality_reward},~\ref{fig:general_alignment_questions},~\ref{fig:general_alignment_reward},~\ref{fig:physical_alignment_reward}, and~\ref{fig:aesthetics_reward}.